\documentclass[twocolumn,final]{IEEEtran}
\usepackage[T1]{fontenc}
\usepackage[latin9]{inputenc}
\usepackage{prettyref}
\usepackage{float}
\usepackage{calc}
\usepackage{amsthm}
\usepackage{amsmath}
\usepackage{amssymb}
\usepackage{graphicx}
\PassOptionsToPackage{normalem}{ulem}
\usepackage{ulem}

\makeatletter

\floatstyle{ruled}
\newfloat{algorithm}{tbp}{loa}
\providecommand{\algorithmname}{Algorithm}
\floatname{algorithm}{\protect\algorithmname}

\theoremstyle{plain}
\newtheorem{thm}{\protect\theoremname}
\theoremstyle{plain}
\newtheorem{prop}[thm]{\protect\propositionname}
\theoremstyle{definition}
\newtheorem{defn}[thm]{\protect\definitionname}
\theoremstyle{plain}
\newtheorem{cor}[thm]{\protect\corollaryname}

\providecommand{\definitionname}{Definition}
\providecommand{\propositionname}{Proposition}
\providecommand{\theoremname}{Theorem}

\title{Local Canonical Correlation Analysis for Nonlinear Common Variables Discovery}
\author{Or~Yair,~\IEEEmembership{Student Member,~IEEE}, Ronen~Talmon,~\IEEEmembership{Member,~IEEE}
\thanks{Or~Yair and Ronen~Talmon are with the Department of Electrical Engineering, Technion -- Israel Institute of Technology, Technion City, Haifa, Israel 32000 (e-mail: oryair@campus.technion.ac.il; ronen@ee.technion.ac.il).}
\thanks{This work was supported by the European Union's Seventh Framework Programme (FP7) under Marie Curie Grant 630657.}}

\usepackage{cite}
\usepackage{hyperref}

\providecommand{\corollaryname}{Corollary}
\providecommand{\definitionname}{Definition}
\providecommand{\propositionname}{Proposition}
\providecommand{\theoremname}{Theorem}

\@ifundefined{showcaptionsetup}{}{%
 \PassOptionsToPackage{caption=false}{subfig}}
\usepackage{subfig}
\makeatother

\providecommand{\corollaryname}{Corollary}
\providecommand{\definitionname}{Definition}
\providecommand{\propositionname}{Proposition}
\providecommand{\theoremname}{Theorem}

\begin{document}
\maketitle




\begin{abstract}
In this paper, we address the problem of hidden common variables discovery
from multimodal data sets of nonlinear high-dimensional observations.
We present a metric based on local applications of canonical correlation
analysis (CCA) and incorporate it in a kernel-based manifold learning
technique. We show that this metric discovers the hidden common variables
underlying the multimodal observations by estimating the Euclidean
distance between them. Our approach can be viewed both as an extension
of CCA to a nonlinear setting as well as an extension of manifold
learning to multiple data sets. Experimental results show that our
method indeed discovers the common variables underlying high-dimensional
nonlinear observations without assuming prior rigid model assumptions. \end{abstract}


\section{Introduction\label{sec:intro}}

\IEEEPARstart{T}{he} need to study and analyze complex systems arises
in many fields. Nowadays, in more and more applications and devices,
many sensors are used to collect and to record multiple channels of
data, a fact that increases the amount of information available to
analyze the state of the system of interest. In such cases, it is
typically insufficient to study each channel separately. Yet, the
ability to gain a deep understanding of the true state of the system
from the overwhelming amount of collected data from multiple (usually
different) sources of information is challenging; it calls for the
development of new technologies and novel ways to observe the system
of interest and to fuse the available information \cite{lahat2015multimodal}.
For example, the study of human physiology in many fields of medicine
is performed by simultaneously monitoring various medical features
through electroencephalography (EEG) signals, electrocardiography
(ECG) signals, respiratory signals, etc. Each type of measurement
carries different and specific information, while our purpose is to
systematically discover an accurate description of the state of the
patient/person.

A commonly-used method that has the ability to reveal correlations
between multiple different sets, which often furthers our understanding
of the system, is the {\em Canonical Correlation Analysis} (CCA)
\cite{hotelling36cca,hardoon2004canonical,bach2005probabilistic}.
CCA is a well known and studied algorithm, where linear projections
maximizing the correlation between the two data sets are constructed.
The main limitation of the CCA algorithm is the inherent restriction
to linear relationships, whereas in medical recordings, for example,
it is unlikely that the collected data carry only linear information
on the human physiological features. To circumvent this linear restriction
and to accommodate nonlinearities, kernel-based extensions of CCA
(KCCA) have been developed, e.g., \cite{lai2000kernel}, which allow
for the discovery of nonlinear relationships between data sets. Indeed,
KCCA has proven to be beneficial in many cases \cite{zheng2006facial,melzer2003appearance,hardoon2007unsupervised}.
Recently, a gamut of work extending CCA based on various combinations
and manipulations of kernels has been presented, e.g., \cite{de2005spectral,wang2012unsupervised,boots2012,Lederman2015,lederman2015alternating}. 

In this paper, we use a different approach using manifold learning
\cite{roweis2000nonlinear,belkin2003laplacian,coifman2006diffusion}.
The core of manifold learning resides in the construction of a kernel
representing affinities between data samples based on pairwise distance
metrics. Such distance metrics define local relationships, which are
then aggregated into a global \emph{nonlinear} representation of the
entire data set. Indeed, in recent studies, various local distance
metrics extending the usage of the prototypical Euclidean metric in
the context of kernel-based manifold learning have been introduced,
e.g. \cite{Singer2008226,talmon2013empirical,Talmon2015138,talmon2015manifold,berry2015local,giannakis2015dynamics,de2010multi,kumar2011co,lin2011multiple,wang2012unsupervised,huang2012affinity,boots2012two,lindenbaum2015multiview,lindenbaum2015learning,michaeli2015nonparametric}.
Along this line of research, our focus in the present work has been
on the construction of a local Riemannian metric for sensor data fusion,
which in turn, can be incorporated in a kernel-based manifold learning
technique.

The contribution of our work is two-fold. First, we present a metric
for discovering the hidden common variables underlying multiple data
sets of observations. Second, we devise a data-driven method based
on this metric that extends manifold learning to multiple data sets
and gives a nonlinear parametrization of the hidden common variables.

Here, we consider the following setting. We assume a system of interest,
observed by two (or more) observation functions. Each observation
captures via a nonlinear and high-dimensional function the system
intrinsic variables. These variables are common to all the observations.
In addition, each observation may introduce additional (noise) variables,
which are specific to each function. In other words, we assume that
our system of interest is monitored via several observations, each
observation, in addition to observing the system itself, observes
additional features which are not related to the system. Consequently,
our focus is on obtaining the common hidden variables which hopefully
represent the true state of the system. Our method includes two main
steps. (i) The construction of a local metric. This metric estimates
the Euclidean distance between any two realizations of the hidden
common variables among nonlinear and high-dimensional data sets. This
is accomplished by using a ``local'' application of CCA, which emphasizes
the common variables underlying the collected data sets while suppressing
the observation-specific features which tend to mask the important
information on the system. We show that the local metric computed
from multiple data sets is a natural extension of a modified {\em
Mahalanobis distance} presented in \cite{Singer2008226,talmon2013empirical,Talmon2015138},
which is computed from only on a single data set. (ii) The usage of
a manifold learning method, Diffusion Maps \cite{coifman2006diffusion},
which recovers a nonlinear global parametrization of the common variables
based on the constructed local metric. Initially, we focus on a setting
with only two data sets, and then, we present an extension of our
method for multiple sets using multi-linear algebra involving tensor
product and tensor decomposition \cite{luo2015tensor,de2000multilinear}.

Experimental results demonstrate that our method is indeed able to
identify the hidden common variables in simulations. In particular,
we present an example of a dynamical system with a definitive underlying
model and demonstrate that without any prior model knowledge, our
method obtains an accurate description of the state of the system
solely from high-dimensional nonlinear observations. Moreover, the
experiments demonstrate the ability of the algorithm to successfully
cope with observations that are only weakly related to the system,
a situation in which we show that KCCA and another recently introduced
method fail.

This paper is organized as follows. In Section \ref{sec:Problem_Formulation}
we formulate the problem. Section \ref{sec:CCA_Background} gives
a brief scientific background presenting CCA and defining the notation
used throughout this paper. In Section \ref{sec:Local} we derive
the local Riemannian metric that estimates the Euclidean distance
between realizations of the hidden common variable, and we present
several results regarding the equality of the estimation. We also
compare the proposed metric to a metric which was recently introduced
in \cite{yair2016multimodal} and show the advantages of the present
one. In Section \ref{sec:Global-Parametrization} we incorporate the
metric into Diffusion Maps and construct a global parametrization
of the common variables. Section \ref{sec:Multi-Observations-Scenario}
presents an extension of our method for the case of multiple (more
than two) data sets. In Section \ref{sec:Experimental-Results}, experimental
results demonstrate the ability to discover an accurate parametrization
of the hidden common variables from multiple data sets of observations
in three different experiments and simulations. Finally, in Section
\ref{sec:Conclusions}, we conclude with several insights and directions
for future work.

\section{Problem Formulation \label{sec:Problem_Formulation}}

We consider a system of interest whose hidden state is governed by
$d_{z}$ isotropic variables $\boldsymbol{z}\sim\mathcal{N}\left(\boldsymbol{\mu},\boldsymbol{I}_{d}\right)$,
$\boldsymbol{z}\in\mathbb{R}^{d_{z}}$. We assume that the hidden
variables $\boldsymbol{z}$ can only be accessed via some observation
functions. In this paper, we focus on the case where the hidden state
of the system is accessed using two (or more) observation functions,
which are possibly nonlinear and are assumed to be locally invertible.
For simplicity, the exposition here focuses on two observation functions.
In Section \ref{sec:Multi-Observations-Scenario}, we present an extension
for more than two. The observations are given by 
\begin{align}
\boldsymbol{x} & =f\left(\boldsymbol{z},\boldsymbol{\epsilon}\right),\qquad\boldsymbol{x}\in\mathbb{R}^{d_{x}},\\
\boldsymbol{y} & =g\left(\boldsymbol{z},\boldsymbol{\eta}\right),\qquad\boldsymbol{y}\in\mathbb{R}^{d_{y}}
\end{align}
where $\boldsymbol{\epsilon}\in\mathbb{R}^{d_{\epsilon}}$ and $\boldsymbol{\eta}\in\mathbb{R}^{d_{\eta}}$
are (hidden) observation-specific variables which depend on the observation
mechanism and are assumed as not related to the system of interest.
The two observation functions can represent, for example, two different
sensors, each introducing additional variables. The probability densities
of the hidden variables $\boldsymbol{\epsilon}$ and \textbf{$\boldsymbol{\eta}$}
are unknown. We assume that the common variables $\boldsymbol{z}$
and the observation-specific variables $\boldsymbol{\epsilon}$ and
$\boldsymbol{\eta}$ are uncorrelated, i.e., $\boldsymbol{\Sigma}_{z\epsilon}=\boldsymbol{\Sigma}_{z\eta}=\boldsymbol{\Sigma}_{\epsilon\eta}=\boldsymbol{0}$,
where $\boldsymbol{\Sigma}_{ab}=\mathbb{E}\left[\boldsymbol{a}\boldsymbol{b}^{T}\right]$.
Finally, we assume that the observations are in higher dimension,
i.e., $d_{z}+d_{\epsilon}\le d_{x},\,d_{z}+d_{\eta}\le d_{y}$.

Given $N$ realizations of the hidden variables, $\left\{ \boldsymbol{z}_{i},\boldsymbol{\epsilon}_{i},\boldsymbol{\eta}_{i}\right\} _{i=1}^{N}$,
we obtain two data sets of observations: 
\begin{align*}
\mathcal{X} & =\left\{ \boldsymbol{x}_{i}\bigg|\,\boldsymbol{x}_{i}=f\left(\boldsymbol{z}_{i},\boldsymbol{\epsilon}_{i}\right)\right\} _{i=1}^{N}\\
\mathcal{Y} & =\left\{ \boldsymbol{y}_{i}\bigg|\,\boldsymbol{y}_{i}=g\left(\boldsymbol{z}_{i},\boldsymbol{\eta}_{i}\right)\right\} _{i=1}^{N}
\end{align*}
Our goal in this paper is to devise a method which builds a parametrization
of the hidden common variables $\boldsymbol{z}$ from the two observation
sets $\mathcal{X}$ and \emph{$\mathcal{Y}$}. The method consists
two main steps. First, a local metric for the hidden variables is
constructed. More specifically, we derive a pairwise metric $D_{ij}$
from the sets $\mathcal{X}$ and $\mathcal{\,Y}$ that corresponds
to the Euclidean distance between the common variables $\boldsymbol{z}$
and neglects the observation-specific variables, $\boldsymbol{\epsilon}$
and $\boldsymbol{\eta}$, i.e., 
\begin{equation}
D_{ij}\approx\left\Vert \boldsymbol{z}_{i}-\boldsymbol{z}_{j}\right\Vert _{2}^{2},\,i,j=1,\dots,N.\label{eq:D_ij}
\end{equation}
Second, manifold learning is applied with a kernel based on the local
metric $D_{ij}$.

\section{Background and Notation\label{sec:CCA_Background}}


Let $v_{x}\triangleq\langle\boldsymbol{p}_{x},\boldsymbol{x}\rangle$
be the inner product between the vector $\boldsymbol{x}$ and a direction
$\boldsymbol{p}_{x}$, defined by $\langle\boldsymbol{p}_{x},\boldsymbol{x}\rangle=\boldsymbol{p}_{x}^{T}\boldsymbol{x}$.
Analogously, let $v_{y}\triangleq\langle\boldsymbol{p}_{y},\boldsymbol{y}\rangle$.
CCA is traditionally applied to two zero mean random vectors $\boldsymbol{x}$
and $\boldsymbol{y}$ and finds the directions that maximize the correlation
between $v_{x}$ and $v_{y}$. The first direction is obtained by
solving the following optimization problem: 
\begin{equation}
\rho^{*}=\max_{\boldsymbol{p}_{x},\boldsymbol{p}_{y}}\rho\left({v}_{x},{v}_{y}\right)\label{eq:cca}
\end{equation}
where the correlation between $v_{x}$ and $v_{y}$ is given by: 
\[
\rho\left(v_{x},v_{y}\right)=\frac{\mathbb{E}\left[v_{x}v_{y}\right]}{\sqrt{\mathbb{E}\left[v_{x}^{2}\right]\mathbb{E}\left[v_{y}^{2}\right]}}
\]
In a similar manner, $d$ directions are obtained iteratively, where
$d\triangleq\min\left(\mbox{rank}\left(\boldsymbol{\Sigma}_{xx}\right),\mbox{rank}\left(\boldsymbol{\Sigma}_{yy}\right)\right)$.
In each iteration, an additional direction is computed by solving
\eqref{eq:cca}, with the restriction that the projection of the random
vector on the current direction is orthogonal to the projections on
the directions attained in previous iterations. Since the correlation
between $v_{x}$ and $v_{y}$ is invariant to (nonzero) scalar multiplication,
we have 
\begin{align}
\rho\left(\alpha v_{x},v_{y}\right) & =\frac{\mathbb{E}\left[\alpha v_{x}\cdot v_{y}\right]}{\sqrt{\mathbb{E}\left[\alpha^{2}v_{x}^{2}\right]\mathbb{E}\left[v_{y}^{2}\right]}}\nonumber \\
 & =\frac{\mathbb{E}\left[v_{x}\cdot v_{y}\right]}{\sqrt{\mathbb{E}\left[v_{x}^{2}\right]\mathbb{E}\left[v_{y}^{2}\right]}}=\rho\left(v_{x},v_{y}\right)
\end{align}
Thus, CCA constrains the projected variable to have a unit variance,
namely $\mathbb{E}\left[v_{x}^{2}\right]=\mathbb{E}\left[v_{y}^{2}\right]=1$.

Using Lagrange multipliers the problem is reduced to an eigenvalue
problem given by 
\begin{equation}
\boldsymbol{\Gamma}\boldsymbol{p}_{x}=\lambda^{2}\boldsymbol{p}_{x}\label{eq:cca_evd}
\end{equation}
where $\boldsymbol{\Gamma}\triangleq\boldsymbol{\Sigma}_{xx}^{-1}\boldsymbol{\Sigma}_{xy}\boldsymbol{\Sigma}_{yy}^{-1}\boldsymbol{\Sigma}_{yx}$
and $\lambda\in\mathbb{R}$ is an unknown scalar. The $d$ right eigenvectors
$\boldsymbol{p}_{x}$ of $\boldsymbol{\Gamma}$ corresponding to the
largest $d$ eigenvalues of $\boldsymbol{\Gamma}$ are solutions of
the optimization problem, where the eigenvalues $\lambda^{2}$ are
the maximal correlations. Thus, the $d$ directions of CCA can be
computed via the eigenvalue decomposition problem \eqref{eq:cca_evd},
circumventing the iterative procedure.

In summary, the application of CCA to two random vectors $\boldsymbol{x}\in\mathbb{R}^{d_{x}}$
and $\boldsymbol{y}\in\mathbb{R}^{d_{y}}$ results in two matrices
$\boldsymbol{P}_{x}\in\mathbb{R}^{d_{x}\times d}$ and $\boldsymbol{P}_{y}\in\mathbb{R}^{d_{y}\times d}$
and a diagonal matrix $\boldsymbol{\Lambda}\in\mathbb{R}^{d\times d}$
, where $\boldsymbol{P}_{x}$ consists of the $d$ directions $\boldsymbol{p}_{x}$,
$\boldsymbol{P}_{y}$ consists of the $d$ directions $\boldsymbol{p}_{y}$,
and $\boldsymbol{\Lambda}$ consists of the $d$ eigenvalues $\lambda^{2}$
on the diagonal. As a result, the random vector $\boldsymbol{v}_{x}=\boldsymbol{P}_{x}^{T}\boldsymbol{x}$
satisfies $\mathbb{\mathbb{E}}\left[\boldsymbol{v}_{x}\boldsymbol{v}_{x}^{T}\right]=\boldsymbol{I}$.
In the same manner, $\boldsymbol{v}_{y}=\boldsymbol{P}_{y}^{T}\boldsymbol{y}$.
In addition, the correlation between the \emph{i}th entry in $\boldsymbol{v}_{x}$
and the \emph{i}th entry in $\boldsymbol{v}_{y}$ is greater or equal
than the correlation between the $(i+1)$th entry. For more details,
see \cite{hotelling36cca}.


\section{Learning the Local Metric \label{sec:Local}}

To obtain a parametrization of the hidden common variables, we construct
a metric that satisfies \eqref{eq:D_ij}, which simultaneously implies
good approximation of the Euclidean distance between any two realizations
of the hidden state variables $\boldsymbol{z}_{i}$ and $\boldsymbol{z}_{j}$,
as well as the attenuation of any effect caused by the observation-specific
variables $\boldsymbol{\epsilon}$ and $\boldsymbol{\eta}$.

\subsection{Linear Case \label{sub:Linear_Case}}

We first describe a special case where $f$ and $g$ are linear functions,
namely: 
\begin{equation}
\boldsymbol{x}=\boldsymbol{J}_{x}\left[\begin{matrix}\boldsymbol{z}\\
\boldsymbol{\epsilon}
\end{matrix}\right],\,\boldsymbol{y}=\boldsymbol{J}_{y}\left[\begin{matrix}\boldsymbol{z}\\
\boldsymbol{\eta}
\end{matrix}\right]\label{eq:Linear_Case}
\end{equation}
where $\boldsymbol{J}_{x}\in\mathbb{R}^{d_{x}\times(d_{z}+d_{\epsilon})}$
and $\boldsymbol{J}_{y}\in\mathbb{R}^{d_{y}\times(d_{z}+d_{\eta})}$.
Note that the assumption $d_{z}+d_{\epsilon}\le d_{x},\,d_{z}+d_{\eta}\le d_{y}$
entails that the set of equations \eqref{eq:Linear_Case} are overdetermined.
By the notation of Section \ref{sec:CCA_Background}, applying CCA
to the random vectors $\boldsymbol{x}$ and \textbf{$\boldsymbol{y}$
}results in the following projection matrices: 
\begin{equation}
\boldsymbol{P}_{x}=\left(\left[\begin{matrix}\boldsymbol{U}_{z} & \boldsymbol{0}\\
\boldsymbol{0} & \boldsymbol{U}_{\epsilon}
\end{matrix}\right]\boldsymbol{J}_{x}^{\dagger}\right)^{T},\,\boldsymbol{P}_{y}=\left(\left[\begin{matrix}\boldsymbol{V}_{z} & \boldsymbol{0}\\
\boldsymbol{0} & \boldsymbol{V}_{\eta}
\end{matrix}\right]\boldsymbol{J}_{y}^{\dagger}\right)^{T}\label{eq:CCA Linear Output 1}
\end{equation}
and with the following correlation matrix: 
\begin{equation}
\boldsymbol{\Lambda}=\left[\begin{matrix}\boldsymbol{I}_{d_{z}} & \boldsymbol{0}\\
\boldsymbol{0} & \boldsymbol{0}
\end{matrix}\right]\qquad\boldsymbol{\Lambda}\in\mathbb{R}^{d\times d}\label{eq:CCA Linear Output 2}
\end{equation}
where $\boldsymbol{U}_{z},\boldsymbol{V}_{z},\boldsymbol{U}_{\epsilon},\boldsymbol{V}_{\eta}$
are arbitrary unitary matrices, and $\boldsymbol{J}_{x}^{\dagger}$
and $\boldsymbol{J}_{y}^{\dagger}$ are the Moore-Penrose pseudoinverse
of $\boldsymbol{J}_{x}$ and $\boldsymbol{J}_{y}$, respectively,
i.e., $\boldsymbol{J}_{x}^{\dagger}\boldsymbol{J}_{x}=\boldsymbol{I}_{\left(d_{z}+d_{\epsilon}\right)\times\left(d_{z}+d_{\epsilon}\right)}$
and $\boldsymbol{J}_{y}^{\dagger}\boldsymbol{J}_{y}=\boldsymbol{I}_{\left(d_{z}+d_{\eta}\right)\times\left(d_{z}+d_{\eta}\right)}$. 
\begin{prop}
\label{prop:Linear_Case} In the linear case, the Euclidean distance
between any two realizations $\boldsymbol{z}_{i}$ and $\boldsymbol{z}_{j}$
of the random variable $\boldsymbol{z}$ is given by: 
\begin{eqnarray*}
\left\Vert \boldsymbol{z}_{i}-\boldsymbol{z}_{j}\right\Vert _{2}^{2} & = & \left(\boldsymbol{x}_{i}-\boldsymbol{x}_{j}\right)^{T}\boldsymbol{P}_{x}\boldsymbol{\Lambda}\boldsymbol{P}_{x}^{T}\left(\boldsymbol{x}_{i}-\boldsymbol{x}_{j}\right)
\end{eqnarray*}
\end{prop}
\begin{IEEEproof}
\begin{align*}
\Delta\boldsymbol{x}^{T}\boldsymbol{P}_{x}\boldsymbol{\Lambda}\boldsymbol{P}_{x}^{T}\Delta\boldsymbol{x} & =\left\Vert \boldsymbol{\Lambda}^{\frac{1}{2}}\boldsymbol{P}_{x}^{T}\Delta\boldsymbol{x}\right\Vert _{2}^{2}\\
 & =\left\Vert \left[\begin{matrix}\boldsymbol{I} & \boldsymbol{0}\\
\boldsymbol{0} & \boldsymbol{0}
\end{matrix}\right]\left[\begin{matrix}\boldsymbol{U}_{z} & \boldsymbol{0}\\
\boldsymbol{0} & \boldsymbol{U}_{\epsilon}
\end{matrix}\right]\boldsymbol{J}_{x}^{\dagger}\boldsymbol{J}_{x}\left[\begin{matrix}\Delta\boldsymbol{z}\\
\Delta\boldsymbol{\epsilon}
\end{matrix}\right]\right\Vert _{2}^{2}\\
 & =\left\Vert \left[\begin{matrix}\boldsymbol{U}_{z} & \boldsymbol{0}\\
\boldsymbol{0} & \boldsymbol{0}
\end{matrix}\right]\left[\begin{matrix}\Delta\boldsymbol{z}\\
\Delta\boldsymbol{\epsilon}
\end{matrix}\right]\right\Vert _{2}^{2}=\left\Vert \Delta\boldsymbol{z}\right\Vert _{2}^{2}
\end{align*}
where $\Delta\boldsymbol{x}=\boldsymbol{x}_{i}-\boldsymbol{x}_{j}$,
$\Delta\boldsymbol{z}=\boldsymbol{z}_{i}-\boldsymbol{z}_{j}$, and
$\Delta\boldsymbol{\epsilon}=\boldsymbol{\epsilon}_{i}-\boldsymbol{\epsilon}_{j}$. 
\end{IEEEproof}
Note that the resulting metric takes into account \emph{only} the
common hidden variables $\boldsymbol{z}$ and filters out the observation-specific
variables $\boldsymbol{\epsilon}$ using the matrix $\boldsymbol{\Lambda}$.
The Euclidean distance between realizations of $\boldsymbol{z}$ can
be expressed in an analogous manner based on realizations of $\boldsymbol{y}$;
one can also take the average of the two metrics using both realizations
of $\boldsymbol{x}$ and $\boldsymbol{y}$.

\subsection{Nonlinear Case\label{sub:Non-Linear-Case}}

In the general case, where $f\left(\boldsymbol{z},\boldsymbol{\epsilon}\right)$
and $g\left(\boldsymbol{z},\boldsymbol{\eta}\right)$ are nonlinear,
we use a linearization approach to obtain a similar result to the
linear case (up to some bounded error). We denote $\boldsymbol{v}_{j}\triangleq\left[\begin{matrix}\boldsymbol{z}_{j}^{T} & \boldsymbol{\epsilon}_{j}^{T}\end{matrix}\right]^{T}$,
such that $\boldsymbol{x}_{j}=f\left(\boldsymbol{v}_{j}\right)$,
and expand $f$ via its Taylor series around $\boldsymbol{v}_{j}$:
\[
\boldsymbol{x}_{i}=\boldsymbol{x}_{j}+\boldsymbol{J}_{x}\left(\boldsymbol{v}_{j}\right)\left[\boldsymbol{v}_{i}-\boldsymbol{v}_{j}\right]+\mathcal{O}\left(\left\Vert \boldsymbol{v}_{i}-\boldsymbol{v}_{j}\right\Vert ^{2}\right)
\]
where $\boldsymbol{J}_{x}\left(\boldsymbol{v}_{j}\right)$ is the
Jacobian of $f$ at $\boldsymbol{v}_{j}$. Reorganizing the expression
above yields: 
\[
f\left(\boldsymbol{v}_{i}\right)=\underbrace{\boldsymbol{x}_{j}-\boldsymbol{J}_{x}\left(\boldsymbol{v}_{j}\right)\boldsymbol{v}_{j}}_{\mbox{Constant}}+\underbrace{\boldsymbol{J}_{x}\left(\boldsymbol{v}_{j}\right)\boldsymbol{v}_{i}}_{\mbox{Linear Part}}+\mathcal{O}\left(\left\Vert \boldsymbol{v}_{i}-\boldsymbol{v}_{j}\right\Vert ^{2}\right).
\]
Thus, when the higher-order terms are negligible, this form is similar
to the linear function considered in Section \ref{sub:Linear_Case}
with an additional constant term. Define the matrices $\boldsymbol{P}_{x}\left(\boldsymbol{x}_{j}\right)$
and $\boldsymbol{\Lambda}\left(\boldsymbol{x}_{j}\right)$ similarly
to \eqref{eq:CCA Linear Output 1} and \eqref{eq:CCA Linear Output 2},
respectively, using $\boldsymbol{J}_{x}\left(\boldsymbol{v}_{j}\right)$
(instead of $\boldsymbol{J}_{x}$) as the linear function. 
\begin{prop}
\label{prop:Non Linear Case} In the nonlinear case, the Euclidean
distance between any two realizations $\boldsymbol{z}_{i}$ and $\boldsymbol{z}_{j}$
of the random variable $\boldsymbol{z}$ is given by: 
\begin{align}
\left\Vert \boldsymbol{z}_{i}-\boldsymbol{z}_{j}\right\Vert _{2}^{2} & =\left(\boldsymbol{x}_{i}-\boldsymbol{x}_{j}\right)^{T}\boldsymbol{A}\left(\bar{\boldsymbol{x}}_{ij}\right)\left(\boldsymbol{x}_{i}-\boldsymbol{x}_{j}\right)\nonumber \\
 & +\mathcal{O}\left(\left\Vert \boldsymbol{x}_{i}-\boldsymbol{x}_{j}\right\Vert ^{4}\right)\label{eq:prop_2}
\end{align}
where $\bar{\boldsymbol{x}}_{ij}\triangleq\left(\boldsymbol{x}_{i}+\boldsymbol{x}_{j}\right)/2$
denotes the middle point, and $\boldsymbol{A}\left(\bar{\boldsymbol{x}}_{ij}\right)\triangleq\boldsymbol{P}_{x}\left(\bar{\boldsymbol{x}}_{ij}\right)\boldsymbol{\Lambda}\left(\bar{\boldsymbol{x}}_{ij}\right)\boldsymbol{P}_{x}^{T}\left(\bar{\boldsymbol{x}}_{ij}\right)$. 
\end{prop}
\begin{IEEEproof}
In the proof of Proposition \prettyref{prop:Linear_Case} we show
that one can write the common variables $\boldsymbol{z}$ up to some
rotation by $\boldsymbol{U}_{z}\boldsymbol{z}=\tilde{\boldsymbol{P}}_{x}^{T}\boldsymbol{x}$,
where $\tilde{\boldsymbol{P}}_{x}$ are the $d_{z}$ leftmost columns
of $\boldsymbol{P}_{x}$. Notice that since $\boldsymbol{U}_{z}$
is unitary, we have $\left\Vert \boldsymbol{z}_{i}-\boldsymbol{z}_{j}\right\Vert _{2}^{2}=\left\Vert \boldsymbol{U}_{z}\boldsymbol{z}_{i}-\boldsymbol{U}_{z}\boldsymbol{z}_{j}\right\Vert _{2}^{2}$.
Thus, since the norm is invariant to rotation, we can recover $\boldsymbol{z}$
up to rotation. With a slight abuse of notation, let $f^{-1}$ denote
the {\em local} inverse function of $f$ restricted to $\boldsymbol{z}$.
The linearization of $f^{-1}$ around the point $\bar{\boldsymbol{x}}_{ij}\triangleq\left(\boldsymbol{x}_{j}+\boldsymbol{x}_{i}\right)/2$
is given by: 
\begin{align*}
\boldsymbol{z}_{i} & =\bar{\boldsymbol{z}}+\tilde{\boldsymbol{P}}_{x}^{T}\left(\bar{\boldsymbol{x}}_{ij}\right)\left[\boldsymbol{x}_{i}-\frac{\boldsymbol{x}_{j}+\boldsymbol{x}_{i}}{2}\right]\\
 & =\bar{\boldsymbol{z}}+\tilde{\boldsymbol{P}}_{x}^{T}\left(\bar{\boldsymbol{x}}_{ij}\right)\left[\frac{\boldsymbol{x}_{i}-\boldsymbol{x}_{j}}{2}\right]
\end{align*}
where $\bar{\boldsymbol{z}}=f^{-1}\left(\bar{\boldsymbol{x}}_{ij}\right)$.
Thus, the Taylor expansion of the $l$th element of $\boldsymbol{z}_{i}$
is given by: 
\begin{align}
\left(z_{i}\right)_{l} & =\left(\bar{\boldsymbol{z}}\right)_{l}+\frac{1}{2}\left(\boldsymbol{p}_{x}^{\left(l\right)}\left(\bar{\boldsymbol{x}}_{ij}\right)\right)^{T}\left(\boldsymbol{x}_{i}-\boldsymbol{x}_{j}\right)+\nonumber \\
 & \frac{1}{8}\left(\boldsymbol{x}_{i}-\boldsymbol{x}_{j}\right)^{T}\boldsymbol{H}_{x}^{\left(l\right)}\left(\bar{\boldsymbol{x}}_{ij}\right)\left(\boldsymbol{x}_{i}-\boldsymbol{x}_{j}\right)+\nonumber \\
 & \mathcal{O}\left(\left\Vert \boldsymbol{x}_{i}-\boldsymbol{x}_{j}\right\Vert ^{3}\right)\label{eq:taylor1}
\end{align}
where $\boldsymbol{p}_{x}^{\left(l\right)}\left(\boldsymbol{x}\right)$
is the \emph{l}-th column of $\boldsymbol{P}_{x}\left(\boldsymbol{x}\right)$,
and $\boldsymbol{H}^{\left(l\right)}\left(\boldsymbol{x}\right)$
is the Hessian of the \emph{l}-th entry of $f^{-1}$. In a similar
way, expanding the $l$-th element of $\boldsymbol{z}_{j}$ around
the same point gives: 
\begin{align}
\left(z_{j}\right)_{l} & =\left(\bar{z}\right)_{l}+\frac{1}{2}\left(\boldsymbol{p}_{x}^{\left(l\right)}\left(\bar{\boldsymbol{x}}_{ij}\right)\right)^{T}\left(\boldsymbol{x}_{j}-\boldsymbol{x}_{i}\right)+\nonumber \\
 & \frac{1}{8}\left(\boldsymbol{x}_{j}-\boldsymbol{x}_{i}\right)^{T}\boldsymbol{H}_{x}^{\left(l\right)}\left(\bar{\boldsymbol{x}}_{ij}\right)\left(\boldsymbol{x}_{j}-\boldsymbol{x}_{i}\right)+\nonumber \\
 & \mathcal{O}\left(\left\Vert \boldsymbol{x}_{j}-\boldsymbol{x}_{i}\right\Vert ^{3}\right)\label{eq:taylor2}
\end{align}
Subtracting \eqref{eq:taylor2} from \eqref{eq:taylor1} yields: 
\begin{eqnarray*}
\left(z_{i}\right)_{l}-\left(z_{j}\right)_{l} & = & \left(\boldsymbol{p}_{x}^{\left(l\right)}\left(\bar{\boldsymbol{x}}_{ij}\right)\right)^{T}\left(\boldsymbol{x}_{i}-\boldsymbol{x}_{j}\right)+\\
 &  & \mathcal{O}\left(\left\Vert \boldsymbol{x}_{i}-\boldsymbol{x}_{j}\right\Vert ^{3}\right)
\end{eqnarray*}
Thus: 
\begin{align*}
\left\Vert \boldsymbol{z}_{i}-\boldsymbol{z}_{k}\right\Vert _{2}^{2} & =\left(\boldsymbol{x}_{i}-\boldsymbol{x}_{j}\right)^{T}\tilde{\boldsymbol{P}}\left(\bar{\boldsymbol{x}}_{ij}\right)\tilde{\boldsymbol{P}}\left(\bar{\boldsymbol{x}}_{ij}\right)^{T}\left(\boldsymbol{x}_{i}-\boldsymbol{x}_{j}\right)\\
 & +\mathcal{O}\left(\left\Vert \boldsymbol{x}_{i}-\boldsymbol{x}_{j}\right\Vert ^{4}\right)\\
 & =\left(\boldsymbol{x}_{i}-\boldsymbol{x}_{j}\right)^{T}\boldsymbol{A}\left(\bar{\boldsymbol{x}}_{ij}\right)\left(\boldsymbol{x}_{i}-\boldsymbol{x}_{j}\right)\\
 & +\mathcal{O}\left(\left\Vert \boldsymbol{x}_{i}-\boldsymbol{x}_{j}\right\Vert ^{4}\right)
\end{align*}
where $\boldsymbol{A}\left(\bar{\boldsymbol{x}}_{ij}\right)\triangleq\boldsymbol{P}_{x}\left(\bar{\boldsymbol{x}}_{ij}\right)\boldsymbol{\Lambda}\left(\bar{\boldsymbol{x}}_{ij}\right)\boldsymbol{P}_{x}^{T}\left(\bar{\boldsymbol{x}}_{ij}\right)$. 
\end{IEEEproof}
We note that similarly to Proposition \ref{prop:Linear_Case}, Proposition
\ref{prop:Non Linear Case} can be analogously formulated based on
realizations of $\boldsymbol{y}$ instead of realizations of $\boldsymbol{x}$.

In \cite{yair2016multimodal} we presented a different way to calculate
the Euclidean distance between two realization: 
\begin{eqnarray*}
\left\Vert \boldsymbol{z}_{i}-\boldsymbol{z}_{j}\right\Vert _{2}^{2} & = & \frac{1}{2}\left(\boldsymbol{x}_{i}-\boldsymbol{x}_{j}\right)^{T}\left[\boldsymbol{A}\left(\boldsymbol{x}_{i}\right)+\boldsymbol{A}\left(\boldsymbol{x}_{j}\right)\right]\left(\boldsymbol{x}_{i}-\boldsymbol{x}_{j}\right)\\
 &  & +\mathcal{O}\left(\left\Vert \boldsymbol{x}_{i}-\boldsymbol{x}_{j}\right\Vert ^{4}\right)
\end{eqnarray*}
where $\boldsymbol{A}\left(\boldsymbol{x}_{i}\right)\triangleq\boldsymbol{P}_{x}\left(\boldsymbol{x}_{i}\right)\boldsymbol{\Lambda}\left(\boldsymbol{x}_{i}\right)\boldsymbol{P}_{x}^{T}\left(\boldsymbol{x}_{i}\right)$.
Yet, both expressions approximate $\left\Vert \boldsymbol{z}_{i}-\boldsymbol{z}_{j}\right\Vert _{2}$
up to the second-order. In Section \ref{sec:Global-Parametrization},
we discuss the advantage of the metric using the middle point $\bar{\boldsymbol{x}}_{ij}$
in terms of computational complexity. In addition, we address the
case where the middle points $(\boldsymbol{x}_{i}+\boldsymbol{x}_{j})/2$
and $(\boldsymbol{y}_{i}+\boldsymbol{y}_{j})/2$ are inaccessible.
In Section \ref{sub:Local-Metric-Comparison}, we compare the proposed
metric using the middle point \eqref{eq:prop_metric} with the metric
proposed in \cite{yair2016multimodal} in a toy problem, which demonstrates
that the computation based on the middle point attains a better estimation
for the Euclidean distance.

\subsection{Implementation \label{sub:Local-CCA}}

Given $\mathcal{X}$, we define a pairwise metric between the $N$
realizations based on Proposition \ref{prop:Non Linear Case}. 
\begin{defn}
Let $D_{ij}$ be the metric between each pair of realizations $\boldsymbol{x}_{i}$
and $\boldsymbol{x}_{j}$ in $\mathcal{X}$, given by: 
\begin{equation}
D_{ij}\triangleq\left(\boldsymbol{x}_{i}-\boldsymbol{x}_{j}\right)^{T}\boldsymbol{A}\left(\bar{\boldsymbol{x}}_{ij}\right)\left(\boldsymbol{x}_{i}-\boldsymbol{x}_{j}\right)\label{eq:prop_metric}
\end{equation}
where $\bar{\boldsymbol{x}}_{ij}\triangleq\frac{\boldsymbol{x}_{i}+\boldsymbol{x}_{j}}{2}$.
We can also define an analogous metric between any two realizations
$\boldsymbol{y}_{i}$ and $\boldsymbol{y}_{j}$ in $\mathcal{Y}$
or define a metric which is the average of the two.

We now describe the computation of the matrices $\boldsymbol{A}\left(\bar{\boldsymbol{x}}_{ij}\right)$
from the sets $\mathcal{X}$ and $\mathcal{Y}$ provided that we have
access to the neighborhoods of the middle points; for the case where
they are inaccessible, see Section \ref{sec:Global-Parametrization}.
For any point $\boldsymbol{x}_{i}$ (including the middle point),
by Proposition \ref{prop:Non Linear Case}, $\boldsymbol{A}\left(\boldsymbol{x}_{i}\right)$
can be computed from the matrices $\boldsymbol{P}_{x}\left(\boldsymbol{x}_{i}\right)$
and $\boldsymbol{\Lambda}\left(\boldsymbol{x}_{i}\right)$. Let $\mathcal{X}_{i}\subset\mathcal{X}$
and $\mathcal{Y}_{i}\subset\mathcal{Y}$ be two subsets of realizations
defining a small neighborhood $\left(\mathcal{X}_{i},\mathcal{Y}_{i}\right)$
around $\left(\boldsymbol{x}_{i},\boldsymbol{y}_{i}\right)$. The
definition of the neighborhoods is application-specific. For example,
for time series data, we could use a time window around each point
to defined its neighbors. Finding the \emph{k nearest neighbors} for
each realization could be another possibility. When considering subsets
$(\mathcal{X}_{i},\mathcal{Y}_{i})$ consisting of only samples $\left(\boldsymbol{x}_{i},\boldsymbol{y}_{i}\right)$
within the neighborhood of $\boldsymbol{x}_{i}$ and $\boldsymbol{y}_{i}$,
then in particular the distance $\left\Vert \boldsymbol{x}_{i}-\boldsymbol{x}_{j}\right\Vert _{2}^{4}$
between any two realizations in the neighborhood is indeed negligible,
and applying CCA to the two sets $\mathcal{X}_{i}$ and $\mathcal{Y}_{i}$
results in the estimation of $\boldsymbol{P}_{x}\left(\boldsymbol{x}_{i}\right)$
and $\boldsymbol{\Lambda}\left(\boldsymbol{x}_{i}\right)$, which
leads to the estimation of $\boldsymbol{A}\left(\boldsymbol{x}_{i}\right)$
as desired.\end{defn}
\begin{prop}
\label{prop:Mahalanobis} In the absence of the observation-specific
variables, the metric $D_{ij}$ can be written as 
\begin{equation}
D_{ij}=\left(\boldsymbol{x}_{i}-\boldsymbol{x}_{j}\right)^{T}\boldsymbol{\Sigma}_{xx}^{-1}\left(\bar{\boldsymbol{x}}_{ij}\right)\left(\boldsymbol{x}_{i}-\boldsymbol{x}_{j}\right)\label{eq:mahalanobis}
\end{equation}
where $\bar{\boldsymbol{x}}_{ij}\triangleq\frac{\boldsymbol{x}_{i}+\boldsymbol{x}_{j}}{2}$
and $\boldsymbol{\Sigma}_{xx}\left(\boldsymbol{x}_{i}\right)$ is
the covariance of the random variable $\boldsymbol{x}$ at the point
$\boldsymbol{x}_{i}$ (noting that the covariance changes from point
to point due to the nonlinearity of the observation function $f$). 
\end{prop}
In other words, when there are no observation-specific variables,
i.e., $\boldsymbol{\epsilon}=\boldsymbol{\eta}=\boldsymbol{0}$, the
metric we build based on local applications of CCA is a modified \emph{Mahalanobis
distance}, which was presented and analyzed in \cite{kushnir2012anisotropic,talmon2013empirical,Talmon2015138}
for the purpose of recovering the intrinsic representation from nonlinear
observation data. 
\begin{IEEEproof}
In the absence of the observation-specific variables, the matrices
$\boldsymbol{\Lambda}\left(\boldsymbol{x}\right)$ become the identity,
namely, $\boldsymbol{\Lambda}\left(\boldsymbol{x}\right)=\boldsymbol{I}$
for all $\boldsymbol{x}$. In addition, a known property of CCA links
between the matrix $\boldsymbol{P}_{x}$ and the covariance matrix
$\boldsymbol{\Sigma}_{xx}$ \cite{EWERBRING198937}: 
\begin{align}
\boldsymbol{P}_{x}^{T}\left(\bar{\boldsymbol{x}}_{ij}\right) & =\boldsymbol{U}^{T}\left(\bar{\boldsymbol{x}}_{ij}\right)\boldsymbol{\Sigma}_{xx}^{-\frac{1}{2}}\left(\bar{\boldsymbol{x}}_{ij}\right)\label{eq:cov_proj}
\end{align}
where $\boldsymbol{U}\left(\boldsymbol{x}\right)$ is a unitary matrix.
We recall that 
\begin{equation}
D_{ij}\triangleq\left(\boldsymbol{x}_{i}-\boldsymbol{x}_{j}\right)^{T}\boldsymbol{A}\left(\bar{\boldsymbol{x}}_{ij}\right)\left(\boldsymbol{x}_{i}-\boldsymbol{x}_{j}\right)\label{eq:Dij}
\end{equation}
and 
\begin{equation}
\boldsymbol{A}\left(\bar{\boldsymbol{x}}_{ij}\right)\triangleq\boldsymbol{P}_{x}\left(\bar{\boldsymbol{x}}_{ij}\right)\boldsymbol{\Lambda}\left(\bar{\boldsymbol{x}}_{ij}\right)\boldsymbol{P}_{x}^{T}\left(\bar{\boldsymbol{x}}_{ij}\right)\label{eq:A}
\end{equation}
Substituting $\boldsymbol{\Lambda}\left(\boldsymbol{x}\right)=\boldsymbol{I}$
and \eqref{eq:cov_proj} into \eqref{eq:A} results in

\begin{eqnarray*}
\boldsymbol{A}\left(\bar{\boldsymbol{x}}_{ij}\right) & = & \boldsymbol{P}_{x}\left(\bar{\boldsymbol{x}}_{ij}\right)\boldsymbol{\Lambda}\left(\bar{\boldsymbol{x}}_{ij}\right)\boldsymbol{P}_{x}^{T}\left(\bar{\boldsymbol{x}}_{ij}\right)\\
 & = & \boldsymbol{\Sigma}_{xx}^{-\frac{1}{2}}\left(\bar{\boldsymbol{x}}_{ij}\right)\boldsymbol{U}\left(\bar{\boldsymbol{x}}_{ij}\right)\boldsymbol{I}\boldsymbol{U}^{T}\left(\bar{\boldsymbol{x}}_{ij}\right)\boldsymbol{\Sigma}_{xx}^{-\frac{1}{2}}\left(\bar{\boldsymbol{x}}_{ij}\right)\\
 & = & \boldsymbol{\Sigma}_{xx}^{-1}\left(\bar{\boldsymbol{x}}_{ij}\right)
\end{eqnarray*}
where we used $\boldsymbol{U}\left(\bar{\boldsymbol{x}}_{ij}\right)\boldsymbol{I}\boldsymbol{U}^{T}\left(\bar{\boldsymbol{x}}_{ij}\right)=\boldsymbol{I}$.
\end{IEEEproof}

\section{Global Parametrization \label{sec:Global-Parametrization}}

In Section \ref{sec:Local}, we proposed a metric that approximates
the Euclidean distance between two realizations. The estimation of
the Euclidean distance is accurate for small distances, whereas the
overall goal is to obtain a global parametrization which corresponds
to the hidden common variables $\boldsymbol{z}$. For this purpose,
i.e., for obtaining a global parametrization from the local metric,
we use a kernel-based manifold learning method, Diffusion Maps \cite{coifman2006diffusion}.
Following common practice, we use a Gaussian kernel $W_{ij}=\exp\left(-D_{ij}/\sigma\right)$,
which emphasizes the notion of locality using the kernel scale $\sigma$:
for $D_{ij}\gg\sigma$, the kernel value $W_{ij}$ is negligible.
Therefore, a proper selection of $\sigma$ entails that only (sufficiently)
small distances $D_{ij}$ are taken into account in the kernel. By
appropriately tuning the value $\sigma$ to correspond to the linear
part of \eqref{eq:prop_2}, $W_{ij}$ accurately represents an affinity
between the common variables, since the higher-order error terms in
\eqref{eq:prop_2} disappear. For more details, see \cite{dsilva2015data}.
The entire method is presented in Algorithm \ref{alg:1}.

\begin{algorithm}[t]
\textbf{\uline{Input}}\textbf{: }Two sets of observations $\mathcal{X}$
and $\mathcal{Y}$.

\textbf{\uline{Output}}\textbf{: }Low dimensional parametrization
of the common variables $\boldsymbol{z}$. 
\begin{enumerate}
\item For each pair of realizations points $\left(\boldsymbol{x}_{i},\boldsymbol{y}_{i}\right),\left(\boldsymbol{x}_{j},\boldsymbol{y}_{j}\right)\in\left(\mathcal{X},\mathcal{Y}\right)$:

\begin{enumerate}
\item Define the middle points $\bar{\boldsymbol{x}}_{ij}\triangleq\frac{1}{2}\left(\boldsymbol{x}_{i}+\boldsymbol{x}_{j}\right)$
and $\bar{\boldsymbol{y}}_{ij}\triangleq\frac{1}{2}\left(\boldsymbol{y}_{i}+\boldsymbol{y}_{j}\right)$. 
\item Construct the subsets $\overline{\mathcal{X}}_{ij},\overline{\mathcal{Y}}_{ij}$
by collecting all pairs $\left(\boldsymbol{x}_{k},\boldsymbol{y}_{k}\right)$
such that $\boldsymbol{x}_{k}$ is in the neighborhood of $\bar{\boldsymbol{x}}_{ij}$
and $\boldsymbol{y}_{k}$ is in the neighborhood of $\bar{\boldsymbol{y}}_{ij}$. 
\item Apply (linear) CCA to the sets $\overline{\mathcal{X}}_{ij},\overline{\mathcal{Y}}_{ij}$
and obtain the matrices $\boldsymbol{P}_{x}\left(\bar{\boldsymbol{x}}_{ij}\right)$
and $\boldsymbol{\Lambda}\left(\bar{\boldsymbol{x}}_{ij}\right)$. 
\item Set $\boldsymbol{A}\left(\bar{\boldsymbol{x}}_{ij}\right)\triangleq\boldsymbol{P}_{x}\left(\bar{\boldsymbol{x}}_{ij}\right)\boldsymbol{\Lambda}\left(\bar{\boldsymbol{x}}_{ij}\right)\boldsymbol{P}_{x}^{T}\left(\bar{\boldsymbol{x}}_{ij}\right)$ 
\item Construct the affinity metric $D_{ij}=\left(\boldsymbol{x}_{i}-\boldsymbol{x}_{j}\right)^{T}\boldsymbol{A}\left(\bar{\boldsymbol{x}}_{ij}\right)\left(\boldsymbol{x}_{i}-\boldsymbol{x}_{j}\right)$. 
\end{enumerate}
\item Apply Diffusion Maps:

\begin{enumerate}
\item Construct the kernel: $W_{ij}=\exp\left(-D_{ij}/\sigma\right)$, where
$\sigma$ is set to the median value of $\{D_{ij}\},\,\forall i,j$. 
\item Normalize the kernel $\boldsymbol{M}=\boldsymbol{\Omega}^{-1}\boldsymbol{W}$,
where $\boldsymbol{\Omega}$ is a diagonal matrix with $\Omega_{ii}=\sum_{j}W_{ij}$ 
\item Compute the eigenvectors and eigenvalues of the matrix $\boldsymbol{M}$,
i.e., $\boldsymbol{M}=\boldsymbol{\Psi}\boldsymbol{S}\boldsymbol{\Psi}^{-1}$. 
\end{enumerate}
\item Form the parametrization of $\boldsymbol{z}_{i},\forall i=1,\ldots,N$
using the $d_{z}$ eigenvectors (the columns of $\boldsymbol{\Psi}$)
associated with the largest $d_{z}$ eigenvalues (without the first
trivial one), i.e., $\left(\Psi_{i1},\ldots,\Psi_{id_{z}}\right)^{T}$
for $i\in1,\dots,N$. 
\end{enumerate}
\protect\protect\caption{Diffusion Maps of Two Datasets With Middle Points\label{alg:1}}
\end{algorithm}

In Step $1$-b, Algorithm \ref{alg:1} assumes that the neighborhoods
of the middle points $\frac{1}{2}\left(\boldsymbol{x}_{i}+\boldsymbol{x}_{j}\right)$
and $\frac{1}{2}\left(\boldsymbol{y}_{i}+\boldsymbol{y}_{j}\right)$
are accessible. In addition, for sets of size $\left|\mathcal{X}\right|=\left|\mathcal{Y}\right|=N$,
Step $1$ is repeated $\frac{N\left(N+1\right)}{2}$ times. This entails
that in Step $1$-d, the matrix $\boldsymbol{A}\left(\bar{\boldsymbol{x}}_{ij}\right)$
is computed for every possible middle point $\bar{\boldsymbol{x}}_{ij}$,
and overall $\frac{N\left(N+1\right)}{2}$ such CCA matrices are computed,
one for each possible pair $\left(\boldsymbol{x}_{i},\boldsymbol{x}_{j}\right)$.

In order to relaxe the above assumption and to reduce the computational
complexity, we present an algorithm based on \cite{kushnir2012anisotropic}.
The more efficient algorithm is presented in Algorithm \prettyref{alg:2},
where the CCA matrices $\boldsymbol{A}\left(\boldsymbol{x}_{i}\right)$
are constructed only for a subset of $L\leq N$ points $\boldsymbol{x}_{i}\in\mathcal{X}_{L}\subseteq\mathcal{X}$
(without the need to directly address the middle point), i.e., $\boldsymbol{A}\left(\boldsymbol{x}_{i}\right)$
is computed only $L$ times. This modification does not affect the
algorithm; Theorem 3.2 presented in \cite{kushnir2012anisotropic}
states that the entries of matrix $\boldsymbol{M}$ calculated in
Step 3 in Algorithm \prettyref{alg:2} are approximations of the entries
of the matrix $\boldsymbol{M}$ calculated in Step 2 in Algorithm
\ref{alg:1}. For more details, see \cite{kushnir2012anisotropic}.
The modification gives rise to two benefits. First, it circumvents
the need to have access to the middle points (and their respective
neighborhoods). Second, in Algorithm \prettyref{alg:2} one can reduce
the computational load by setting $\mathcal{X}_{L}\subset\mathcal{X}$
and then by calculating $\boldsymbol{A}\left(\boldsymbol{x}_{i}\right)$
only at $L<N$ different points.

\begin{algorithm}[t]
\textbf{\uline{Input}}\textbf{: }Two sets of observations $\mathcal{X}$
and $\mathcal{Y}$.

\textbf{\uline{Output}}\textbf{: }Low dimensional parametrization
of the common variables $\boldsymbol{z}$. 
\begin{enumerate}
\item Construct some subsets $\mathcal{X}_{L}\subseteq\mathcal{X}$ and
$\mathcal{Y}_{L}\subseteq\mathcal{Y}$ with $\left|\mathcal{X}_{L}\right|=\left|\mathcal{Y}_{L}\right|=L\leq N$
such that $\boldsymbol{x}_{i}\in\mathcal{X}_{L}$ iff \emph{$\boldsymbol{y}_{i}\in\mathcal{Y}_{L}$} 
\item For each pair of points $\left(\boldsymbol{x}_{i},\boldsymbol{y}_{i}\right)\in\left(\mathcal{X}_{L},\mathcal{Y}_{L}\right)$:

\begin{enumerate}
\item Construct the subsets $\mathcal{X}_{i},\mathcal{Y}_{i}$ by choosing
all pairs $\left(\boldsymbol{x}_{j},\boldsymbol{y}_{j}\right)$ such
that $\boldsymbol{x}_{j}$ is in the neighborhood of $\boldsymbol{x}_{i}$
and $\boldsymbol{y}_{j}$ is in the neighborhood of $\boldsymbol{y}_{i}$. 
\item Apply (linear) CCA to the sets $\mathcal{X}_{i},\mathcal{Y}_{i}$
and obtain the matrices $\boldsymbol{P}_{x}\left(\boldsymbol{x}_{i}\right)$
and $\boldsymbol{\Lambda}\left(\boldsymbol{x}_{i}\right)$. 
\item Set $\boldsymbol{A}\left(\boldsymbol{x}_{i}\right)\triangleq\boldsymbol{P}_{x}\left(\boldsymbol{x}_{i}\right)\boldsymbol{\Lambda}\left(\boldsymbol{x}_{i}\right)\boldsymbol{P}_{x}^{T}\left(\boldsymbol{x}_{i}\right)$ 
\end{enumerate}
\item For each two observations $\boldsymbol{x}_{i}\in\mathcal{X}_{L}$
and $\boldsymbol{x}_{j}\in\mathcal{X}$,\\
 construct the affinity metric $\widetilde{D}\in\mathbb{R}^{L\times N}$
according to 
\[
\widetilde{D}_{ij}=\left(\boldsymbol{x}_{i}-\boldsymbol{x}_{j}\right)^{T}\boldsymbol{A}\left(\boldsymbol{x}_{i}\right)\left(\boldsymbol{x}_{i}-\boldsymbol{x}_{j}\right)
\]

\item Apply Diffusion Maps:

\begin{enumerate}
\item Construct the kernel: $W_{ij}=\exp\left(-\widetilde{D}_{ij}/\sigma\right)$,
where $\sigma$ is set to the median value of $\{\widetilde{D}_{i,j}\},\,\forall i,j$. 
\item Normalize the kernel $\boldsymbol{M}=\boldsymbol{\Omega}^{-\frac{1}{2}}\boldsymbol{W}^{T}\boldsymbol{W}\boldsymbol{\Omega}^{-\frac{1}{2}}$,
where $\boldsymbol{\Omega}$ is a diagonal matrix with $\Omega_{ii}=\sum_{j}\left(W^{T}W\right)_{ij}$ 
\item Compute the eigenvectors and eigenvalues of the matrix $\boldsymbol{M}$,
i.e., $\boldsymbol{M}=\boldsymbol{\Psi}\boldsymbol{S}\boldsymbol{\Psi}^{-1}$. 
\end{enumerate}
\item Form the parametrization of $\boldsymbol{z}_{i},\forall i=1,\ldots,N$
using the $d_{z}$ eigenvectors (the columns of $\boldsymbol{\Psi}$)
associated with the largest $d_{z}$ eigenvalues (without the first
trivial one), i.e., $\left(\Psi_{i1},\ldots,\Psi_{id_{z}}\right)^{T}$
for $i\in1,\dots,N$. 
\end{enumerate}
\protect\protect\caption{Diffusion Maps of Two Datasets Without Middle Points\label{alg:2}}
\end{algorithm}

\section{Multiple Observation Scenario \label{sec:Multi-Observations-Scenario}}

The proposed method can be extended to the case where there are more
than two sets of observations. The core of the proposed method relies
on the local metric \prettyref{eq:prop_metric}. As described in \prettyref{sec:Local},
this metric requires the computation of the matrix $\boldsymbol{A}$
(at the middle points in Algorithm 1 or at the observations in Algorithm
2). In either case, this matrix is computed based on the canonical
directions extracted by a local application of CCA to two sets of
observations from two (possibly different) observation functions.
Consequently, the extension to more than two sets involves an extension
of the local CCA application that enables to compute the canonical
directions from more than two sets of observations. Once such canonical
directions are identified, $\boldsymbol{A}$ can be constructed analogously,
and the remainder of the algorithm remains unchanged.

Therefore, this multiple observations case requires a suitable alternative
to CCA, which is not restricted to two sets. Here, we exploit the
method presented in \cite{luo2015tensor}, which extends CCA for the
case of more than two data sets. As mentioned in \cite{luo2015tensor},
this method is limited to finding only the dominant canonical direction
for each observation, since finding multiple directions that satisfy
the orthogonality constraint is still an open problem \cite{de2000multilinear}.

In the remainder of this section, we extend our notation to support
multiple observations. Then, we briefly describe the (linear) Tensor
CCA (TCCA) method based on \cite{luo2015tensor} using multi-linear
algebra, i.e. using tensor products and tensor decompositions. Finally,
we present an algorithm for the general multimodal scenario, which
extends the algorithm presented in Section \ref{sec:Global-Parametrization}
for more than two sets of observations.

Let $f^{\left(k\right)}$ denote the $k$th observation function,
i.e., $\boldsymbol{x}^{\left(k\right)}=f^{\left(k\right)}\left(\boldsymbol{z},\boldsymbol{\epsilon}^{\left(k\right)}\right)$,
where $\boldsymbol{\epsilon}^{\left(k\right)}$ is a \emph{k}th observation-specific
variable. Let $\mathcal{X}^{\left(k\right)}=\left\{ \boldsymbol{x}_{i}^{\left(k\right)}\right\} _{i=1}^{N}$
denote the \emph{k}th set of observations, where $1\leq k\leq K$.

As mentioned above, extending the derivation of the local metric presented
in Section \ref{sec:Global-Parametrization} for $K>2$ observation
sets requires the use of tensors instead of matrices. The notation
that is used throughout this section is as follows. 
\begin{defn}
The $k$th $\left(1\leq k\leq K\right)$ mode product between a \emph{K}th
order tensor $\mathcal{T}\in\mathbb{R}^{d_{1}\times d_{2}\times\dots\times d_{K}}$
and a matrix $M\in\mathbb{R}^{d_{k}\times D}$ is defined by 
\begin{align*}
\mathcal{K}\left(m_{1},\ldots,m_{k-1},n,m_{k+1},\dots m_{K}\right)\\
=\sum_{m_{k}=1}^{d_{k}}\mathcal{T}\left(m_{1},\ldots,m_{K}\right)M\left(m_{k},n\right).
\end{align*}
In matrix form this product can be expressed by $\mathcal{K}=\mathcal{T}\times_{k}M$,
where 
\[
\mathcal{K}\in\mathbb{R}^{d_{1}\times d_{2}\times\dots d_{k-1}\times D\times d_{k+1}\times\cdots d_{K}}
\]
In a similar manner, we define by 
\[
\mathcal{K}=\mathcal{T}\times_{1}M_{1}\times_{2}M_{2}\dots\times_{K}M_{K}
\]
the product of $\mathcal{T}$ with a sequence of matrices $M_{k}\in\mathbb{R}^{d_{k}\times D_{k}}$
for $1\leq k\leq K$. Note that in this case $\mathcal{K}\in\mathbb{R}^{D_{1}\times D_{2}\times\dots\times D_{K}}$. 
\end{defn}

\begin{defn}
Given a \emph{K}th order tensor $\mathcal{T}_{1}\in\mathbb{R}^{d_{1}\times d_{2}\times\dots\times d_{K}}$,
and a \emph{J}th order tensor $\mathcal{T}_{2}\in\mathbb{R}^{D_{1}\times D_{2}\times\dots\times D_{J}}$,
their outer product $\mathcal{T}_{1}\otimes\mathcal{T}_{2}$ is a
$(K+J)$th order tensor $\mathcal{K}=\mathcal{T}_{1}\otimes\mathcal{T}_{2}\in\mathbb{R}^{d_{1}\times\cdots\times d_{K}\times D_{1}\times\cdots\times D_{J}}$
which holds: 
\begin{align*}
\mathcal{K} & \left(m_{1},m_{2},\dots,m_{K},m_{K+1},\dots,m_{K+J}\right)\\
 & =\mathcal{T}_{1}\left(m_{1},\dots m_{K}\right)\mathcal{T}_{2}\left(m_{K+1},\dots,m_{K+J}\right)
\end{align*}

\end{defn}
In Section \ref{sub:Non-Linear-Case}, we estimate the Euclidean distance
$\left\Vert \boldsymbol{z}_{i}-\boldsymbol{z}_{j}\right\Vert _{2}$
by projecting the observations from each set on the respective canonical
directions obtained by a local application of CCA. In the case of
multiple sets, we aim to find the generalized canonical directions
which maximize the correlation of observations $\boldsymbol{x}^{\left(k\right)}$
from all sets $k=1,\ldots,K$. Assuming zero mean random variables
for simplicity, the corresponding optimization problem can be written
as follows: 
\begin{align}
\arg\max_{\left\{ \boldsymbol{p}^{\left(k\right)}\right\} _{k=1}^{K}}\rho\left(v_{1},v_{2},...v_{K}\right)\nonumber \\
\mbox{s.t.\,}\mathbb{E}\left[v_{k}^{2}\right]=1,\ k\in\left\{ 1,\dots,K\right\} \label{eq:tcca_opt_prob}
\end{align}
where $v_{k}\triangleq\langle\boldsymbol{p}^{\left(k\right)},\boldsymbol{x}^{(k)}\rangle$
and $\rho\left(v_{1},v_{2},...v_{K}\right)=\mathbb{E}\left[v_{1}v_{2}\cdot\dots\cdot v_{k}\right]$.

A summary of the algorithm presented in \cite{luo2015tensor} for
obtaining the generalized canonical directions is outlined in Algorithm
\ref{alg:Linear-TCCA}. Note that in Step 4, Algorithm \ref{alg:Linear-TCCA}
uses a low-rank tensor decomposition to solve the optimization problem
\eqref{eq:tcca_opt_prob}. To compute this decomposition, one can
use the alternating least squares (ALS) algorithm \cite{comon2009tensor}.

We repeat the same steps done in Section \ref{sub:Non-Linear-Case}
to obtain the generalized canonical direction $\boldsymbol{p}^{\left(k\right)}$
at the point $\boldsymbol{x}_{i}^{\left(k\right)}$, namely, $\boldsymbol{p}^{\left(k\right)}\left(\boldsymbol{x}_{i}^{\left(k\right)}\right)$.
In other words, we apply Algorithm \ref{alg:Linear-TCCA} only to
the neighborhood of the \emph{i}th realizations, $\mathcal{X}_{i}^{(k)}$.
In addition, by repeating the same steps as in the proof of Proposition
\ref{prop:Non Linear Case}, we arrive to the following result. 
\begin{cor}
\label{cor:TCCA}In the multiple observation case, the Euclidean distance
between any two (scalar) realizations $z_{i}$ and $z_{j}$ of the
random variable $z$ is given by: 
\begin{align*}
\left\Vert z_{i}-z_{j}\right\Vert _{2}^{2} & =\left(\boldsymbol{x}_{i}^{\left(1\right)}-\boldsymbol{x}_{j}^{\left(1\right)}\right)^{T}\boldsymbol{A}\left(\bar{\boldsymbol{x}}_{ij}^{\left(1\right)}\right)\left(\boldsymbol{x}_{i}^{\left(1\right)}-\boldsymbol{x}_{j}^{\left(1\right)}\right)\\
 & +\mathcal{O}\left(\left\Vert \boldsymbol{x}_{i}^{\left(1\right)}-\boldsymbol{x}_{j}^{\left(1\right)}\right\Vert ^{4}\right)
\end{align*}
where $\bar{\boldsymbol{x}}_{ij}^{\left(1\right)}\triangleq\left(\boldsymbol{x}_{i}^{\left(1\right)}+\boldsymbol{x}_{j}^{\left(1\right)}\right)/2$
and $\boldsymbol{A}\left(\bar{\boldsymbol{x}}_{ij}\right)\triangleq\left(\boldsymbol{p}^{\left(1\right)}\left(\bar{\boldsymbol{x}}_{ij}^{\left(1\right)}\right)\right)\left(\boldsymbol{p}^{\left(1\right)}\left(\bar{\boldsymbol{x}}_{ij}^{\left(1\right)}\right)\right)^{T}$. 
\end{cor}
Note that the common variable $z$ in this case is restricted to be
a scalar, due to the limitation of the TCCA algorithm. We note that
similarly to Proposition \ref{prop:Non Linear Case}, Corollary \ref{cor:TCCA}
can be analogously formulated based on realizations of $\boldsymbol{x}^{\left(k\right)}$
(instead of $\boldsymbol{x}^{\left(1\right)}$) for any $1\leq k\leq K$.

\begin{algorithm}
\textbf{\uline{Input}}\textbf{:} $K$ sets of observations $\mathcal{X}^{\left(k\right)}$,
$1\leq k\leq K$

\textbf{\uline{Output}}\textbf{:} The canonical directions $\boldsymbol{p}^{\left(1\right)},\boldsymbol{p}^{\left(2\right)},\dots\boldsymbol{p}^{\left(K\right)}$ 
\begin{enumerate}
\item For each set $\mathcal{X}^{\left(k\right)}=\left\{ \boldsymbol{x}_{i}^{\left(k\right)}\right\} _{i=1}^{N}$,
compute the covariance matrix $\boldsymbol{\Sigma}_{kk}$. 
\item Compute the covariance tensor 
\[
\mathcal{C}_{12\dots K}=\frac{1}{N}\sum_{i=1}^{N}\boldsymbol{x}_{i}^{\left(1\right)}\otimes\boldsymbol{x}_{i}^{\left(2\right)}\otimes\dots\otimes\boldsymbol{x}_{i}^{\left(K\right)}
\]

\item Compute: 
\[
\mathcal{T}=\mathcal{C}_{12\dots K}\times_{1}\Sigma_{11}^{-\frac{1}{2}}\times_{2}\Sigma_{22}^{-\frac{1}{2}}\times\dots\times_{K}\Sigma_{KK}^{-\frac{1}{2}}
\]

\item Apply a rank-$1$ tensor approximation to $\mathcal{T}$ by solving:
\[
\arg\min_{\rho,\left\{ \boldsymbol{p}^{\left(k\right)}\right\} _{k=1}^{K}}\left\Vert \mathcal{T}-\rho\boldsymbol{p}^{\left(1\right)}\otimes\boldsymbol{p}^{\left(2\right)}\otimes\dots\otimes\boldsymbol{p}^{\left(K\right)}\right\Vert _{F}
\]
where $\left\Vert \cdot\right\Vert _{F}$ is the Frobenius norm, and
obtain the canonical directions. 
\end{enumerate}
\protect\protect\protect\protect\protect\caption{Linear TCCA \label{alg:Linear-TCCA}}
\end{algorithm}

Obtaining the global metric from the local metric is achieved similarly
to the case where $K=2$ and is described in Section \ref{sub:Non-Linear-Case}.
Here as well, we use Diffusion Maps with a Gaussian kernel. The overall
algorithm for the multimodal case ($K>2$) is presented in Algorithm
\ref{alg:full_TCCA}.

\begin{algorithm}[tbh]
\textbf{\uline{Input}}\textbf{: }$K$ sets of observations $\mathcal{X}^{\left(k\right)}$,
$1\leq k\leq K$

\textbf{\uline{Output}}\textbf{: }Low dimensional parametrization
of the common variable $z$. 
\begin{enumerate}
\item For each sample $\left(\boldsymbol{x}_{i}^{\left(1\right)},\boldsymbol{x}_{i}^{\left(2\right)},\dots,\boldsymbol{x}_{i}^{\left(K\right)}\right)\in\left(\mathcal{X}^{\left(1\right)},\mathcal{X}^{\left(2\right)},\dots,\mathcal{X}^{\left(K\right)}\right)$:

\begin{enumerate}
\item Construct the subsets $\left(\mathcal{X}_{i}^{\left(1\right)},\mathcal{X}_{i}^{\left(2\right)},\dots,\mathcal{X}_{i}^{\left(K\right)}\right)$
by choosing all samples $\left(\boldsymbol{x}_{j}^{\left(1\right)},\boldsymbol{x}_{j}^{\left(2\right)},\dots,\boldsymbol{x}_{j}^{\left(K\right)}\right)$
such that the \emph{k}th coordinate $\boldsymbol{x}_{j}^{\left(k\right)}$
is in the neighborhood of $\boldsymbol{x}_{i}^{\left(k\right)}$. 
\item Apply (linear) TCCA to the sets $\left(\mathcal{X}_{i}^{\left(1\right)},\mathcal{X}_{i}^{\left(2\right)},\dots,\mathcal{X}_{i}^{\left(K\right)}\right)$
and obtain the vector $\boldsymbol{p}^{\left(k\right)}\left(\boldsymbol{x}_{i}\right)$. 
\end{enumerate}
\item For each two observations $\boldsymbol{x}_{i}^{\left(k\right)},\boldsymbol{x}_{j}^{\left(k\right)}\in\mathcal{X}^{\left(k\right)}$,
construct the affinity metric $\widetilde{D}_{ij}$ according to 
\[
\widetilde{D}_{ij}=\left(\boldsymbol{x}_{i}^{\left(k\right)}-\boldsymbol{x}_{j}^{\left(k\right)}\right)^{T}\boldsymbol{A}^{\left(k\right)}\left(\boldsymbol{x}_{i}^{\left(k\right)}\right)\left(\boldsymbol{x}_{i}^{\left(k\right)}-\boldsymbol{x}_{j}^{\left(k\right)}\right)
\]
where $\boldsymbol{A}^{\left(k\right)}\left(\boldsymbol{x}_{i}^{\left(k\right)}\right)\triangleq\left(\boldsymbol{p}^{\left(k\right)}\left(\boldsymbol{x}_{i}^{\left(k\right)}\right)\right)\left(\boldsymbol{p}^{\left(k\right)}\left(\boldsymbol{x}_{i}^{\left(k\right)}\right)\right)^{T}$. 
\item Apply Diffusion Maps:

\begin{enumerate}
\item Construct the kernel: $W_{ij}=\exp\left(-\widetilde{D}_{ij}/\sigma\right)$,
where $\sigma$ is set to the median values $\widetilde{D}_{i,j},\,\forall i,j$. 
\item Normalize the kernel $\boldsymbol{M}=\boldsymbol{\Omega}^{-\frac{1}{2}}\boldsymbol{W}^{T}\boldsymbol{W}\boldsymbol{\Omega}^{-\frac{1}{2}}$,
where $\boldsymbol{\Omega}$ is a diagonal matrix with $\Omega_{ii}=\sum_{j}\left({W}^{T}{W}\right)_{ij}$ 
\item Compute the eigenvectors and eigenvalues of the matrix $\boldsymbol{M}$,
i.e., $\boldsymbol{M}=\boldsymbol{\Psi}\boldsymbol{S}\boldsymbol{\Psi}^{-1}$. 
\end{enumerate}
\item Form the parametrization of $z_{i},\forall i=1,\ldots,N$ using the
the eigenvector (the left most column of $\boldsymbol{\Psi}$) associated
with the largest eigenvalue (excluding the trivial eigenvector). 
\end{enumerate}
\protect\protect\caption{Diffusion Maps of $K$ Datasets \label{alg:full_TCCA}}
\end{algorithm}

\section{Experimental Results \label{sec:Experimental-Results}}

\subsection{Local Metric Comparison \label{sub:Local-Metric-Comparison}}

\begin{figure}
\centering{}\subfloat[]{\protect

\protect\begin{centering}
\protect\protect\includegraphics[width=0.45\columnwidth]{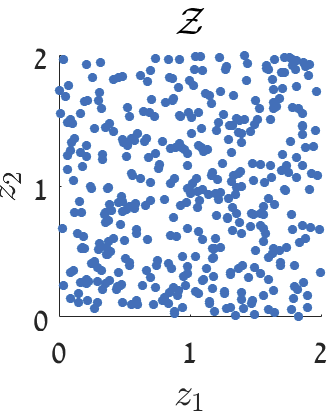}\protect \protect
\par\end{centering}

}\subfloat[]{\protect\centering{}\protect\protect\includegraphics[width=0.45\columnwidth]{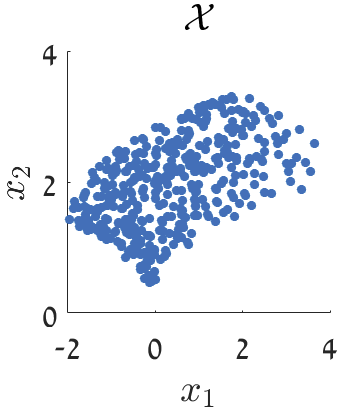}\protect

}

\protect\protect\caption{(a) The hidden random variables $\mathcal{Z}=\left\{ \boldsymbol{z}_{i}\right\} _{i=1}^{N}$.
(b) The observations $\mathcal{X}=\left\{ \boldsymbol{x}_{i}\right\} _{i=1}^{N}$.}
\label{fig:Z_And_X_Sets} 
\end{figure}

We generate $N=400$ realizations $\left\{ \boldsymbol{z}_{i}\right\} _{i=1}^{N}$
of a two dimensional random variable with uniform distribution in
the $\left[0,2\right]^{2}$ square. To compare only the metric estimation,
we use an observation with no observation-specific variables. According
to Proposition \ref{prop:Mahalanobis}, only a single observation
is needed, and we can compare between $D_{ij}$ as defined in \eqref{eq:Dij}
and the metric defined in \cite{yair2016multimodal}, which we denote
here as $Q_{ij}$, namely: 
\[
Q_{ij}=\frac{1}{2}\left(\boldsymbol{x}_{i}-\boldsymbol{x}_{j}\right)^{T}\left[\boldsymbol{A}\left(\boldsymbol{x}_{i}\right)+\boldsymbol{A}\left(\boldsymbol{x}_{j}\right)\right]\left(\boldsymbol{x}_{i}-\boldsymbol{x}_{j}\right).
\]
By simulating the following nonlinear observation function: 
\[
\boldsymbol{x}=f\left(\boldsymbol{z}\right)=\left[\begin{matrix}z_{1}^{2}-z_{2}\\
z_{1}+\sqrt{z_{2}}
\end{matrix}\right]
\]
we obtain a set of $N=400$ observations $\mathcal{X}=\left\{ \boldsymbol{x}_{i}\right\} _{i=1}^{N}$.
Figure \ref{fig:Z_And_X_Sets} depicts (a) the hidden variables $\mathcal{Z}=\left\{ \boldsymbol{z}_{i}\right\} $
and (b) the observations $\mathcal{X}=\left\{ \boldsymbol{x}_{i}\right\} $.

\begin{figure}[t]
\subfloat[]{\centering \protect\includegraphics[width=0.45\columnwidth]{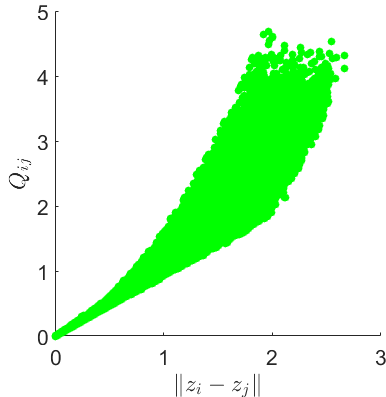}

}\hfill{}\subfloat[]{\protect\includegraphics[width=0.45\columnwidth]{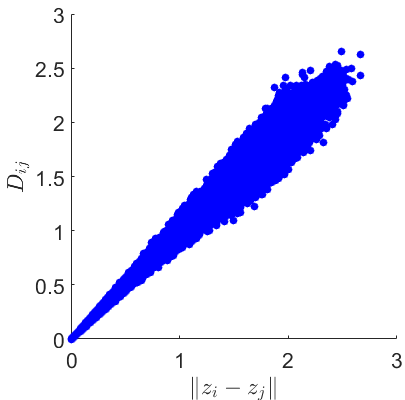}

}

\protect\caption{Comparison of Euclidean metric estimation. (a) The metric estimation
based on \cite{yair2016multimodal}. (b) The metric estimation based
on the middle point \eqref{eq:Dij}.}
\label{fig:Comparison-of-Euclidean} 
\end{figure}

Figure~\ref{fig:Comparison-of-Euclidean} shows the estimated metric
as a function of the true metric. In Figure~\ref{fig:Comparison-of-Euclidean}(a),
we plot the estimated metric based on \cite{yair2016multimodal},
and in Figure~\ref{fig:Comparison-of-Euclidean}(b) we plot the estimated
metric based on \prettyref{eq:Dij}. We can see that for small Euclidean
distances (small values on the x-axis), both estimated metrics are
accurate. For large Euclidean distances, the estimated metric based
on the middle point maintains a linear correlation with the true distance,
whereas the estimated metric proposed in \cite{yair2016multimodal}
exhibits large error.

\begin{figure}[t]
\begin{centering}
\includegraphics[width=0.35\columnwidth]{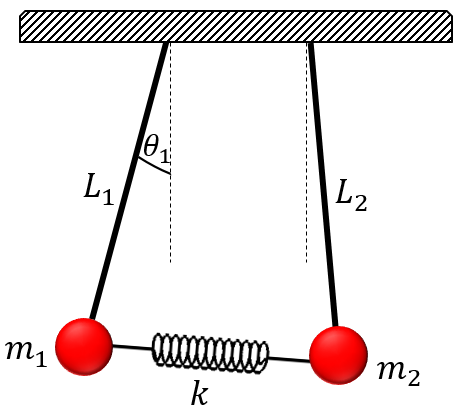} 
\par\end{centering}

\protect\protect\protect\caption{Illustration of the setup of the coupled pendulums system.}
\label{fig:Coupled_Pendulum} 
\end{figure}

\subsection{Coupled Pendulum}

this experiment we simulate a coupled pendulum model. This model consists
of two simple pendulums with lengths $L_{1}$ and $L_{2}$ and masses
$m_{1}$ and $m_{2}$, which are connected by a spring as shown in
Figure \ref{fig:Coupled_Pendulum}. For simplicity we set the same
length and the same mass for both pendulums, namely $L_{1}=L_{2}=L$
and $m_{1}=m_{2}=m$. Let $u^{(i)}\left(t\right)$ and $w^{(i)}\left(t\right)$
denote the horizontal position and vertical position of the $i$th
pendulum, respectively. Note that a close-form expression for the
positions cannot be derived for the general case. Yet, in the case
of small perturbations around the equilibrium point, we can consider
a linear regime. Accordingly, let $\theta^{(i)}(t)=\arctan\left(u^{(i)}(t)/w^{(i)}(t)\right)$
be the angle between the $i$th pendulum and the vertical axis, and
assume that $w^{(1)}\left(t\right)=w^{(2)}\left(t\right)=-L$, and
$\sin\left(\theta^{(i)}\right)\approx\theta^{(i)}$. The ordinary
differential equation (ODE) representing the horizontal position under
the linear regime is given by: 
\begin{align}
\begin{cases}
m\ddot{u}^{(1)}=-\frac{mg}{L}u^{(1)}-k\left(u^{(2)}-u^{(1)}\right)\\
m\ddot{u}^{(2)}=-\frac{mg}{L}u^{(2)}+k\left(u^{(2)}-u^{(1)}\right)
\end{cases}\label{eq:ODE}
\end{align}
where $\ddot{u}$ is the second derivative of $u$, $g$ is the gravity
of earth, and $k$ is the spring constant. For the following initial
conditions: 
\[
\dot{u}_{1}\left(0\right)=0,u_{1}\left(0\right)=\delta,\dot{u}_{2}\left(0\right)=0,u_{2}\left(0\right)=0
\]
where $0<\delta\in\mathbb{R}$ is assumed to be sufficiently small
to satisfy the linear regime, the closed-form solution of the ODE
\eqref{eq:ODE} is given by: 
\begin{equation}
\begin{cases}
u^{(1)}\left(t\right)=\frac{1}{2}\delta\cos\left(\omega_{1}t\right)+\frac{1}{2}\delta\cos\left(\omega_{2}t\right)\\
u^{(2)}\left(t\right)=\frac{1}{2}\delta\cos\left(\omega_{1}t\right)-\frac{1}{2}\delta\cos\left(\omega_{2}t\right)
\end{cases}\label{eq:ODE_Solution}
\end{equation}
where 
\[
\omega_{1}=\sqrt{\frac{g}{L}},\ \ \omega_{2}=\sqrt{\frac{g}{L}+\frac{2k}{m}}
\]

This example suits our purposes, since the horizontal displacement
of each pendulum $u^{(i)}\left(t\right)$ is a linear combination
of two harmonic motions with the {\em common} frequencies $\omega_{1}$
and $\omega_{2}$. In other words, the horizontal displacement of
each pendulum can be viewed as a different observation of the same
common harmonic motion.

\begin{figure}[t]
\includegraphics[width=0.95\columnwidth]{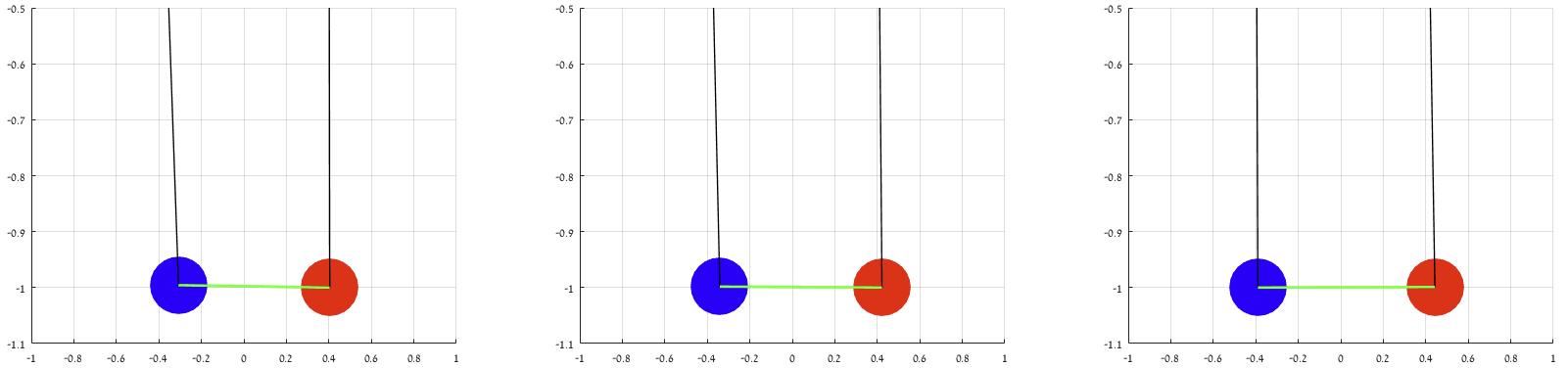} \protect\protect\protect\caption{An example of $3$ snapshots of the coupled pendulum system.}
\label{fig:Entire_Clean_Movie} 
\end{figure}

To further demonstrate the power of our method, we assume that we
do not have direct access to the horizontal displacement. Instead,
we generate movies of the motion of the coupled pendulum in the linear
regime. Consequently, on the one hand, we have a definitive ground
truth described by the solution of the ODE of the system (the harmonic
motion with the two frequencies $\omega_{1}$ and $\omega_{2}$).
On the other hand, we only have access to high-dimensional nonlinear
observations of the system, and we do not assume any prior model knowledge.
Three snapshots of the entire system are displayed in Figure \ref{fig:Entire_Clean_Movie}.

This model is used to test our method in two scenarios. In the first
scenario, we generate two movies of the two pendulums without any
other features. In the second scenario, we generate two movies of
the two pendulums, where each movie also contains an additional pendulum
that represents an observation-specific geometric noise.

\subsubsection{Case I -- Coupled pendulum}

\label{sub:Simulation_Case_I}

\begin{figure}[t]
\begin{centering}
\subfloat[]{\protect

\protect

\protect\begin{centering}
\protect\protect\protect\includegraphics[width=0.95\columnwidth]{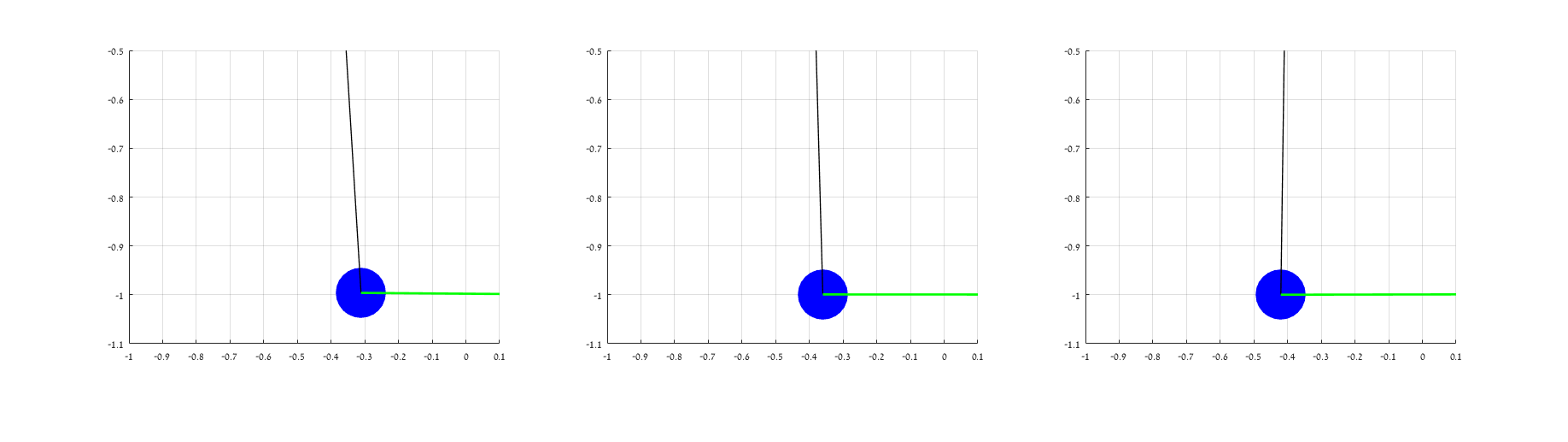}
\protect \protect \protect
\par\end{centering}

}
\par\end{centering}

\begin{centering}
\subfloat[]{\protect

\protect

\protect\begin{centering}
\protect\protect\protect\includegraphics[width=0.95\columnwidth]{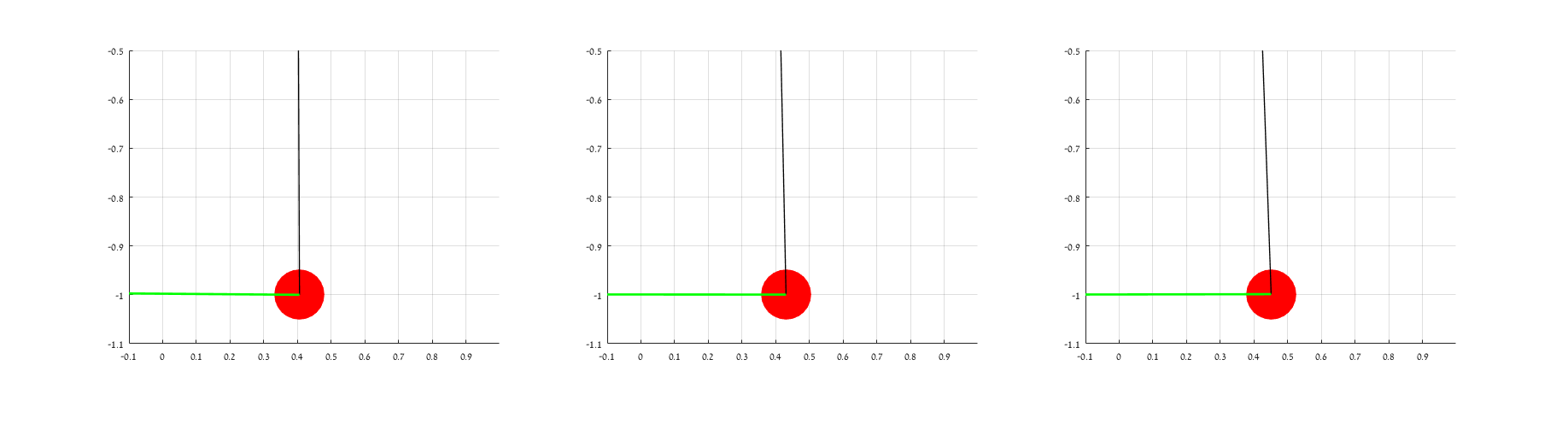}
\protect \protect \protect
\par\end{centering}

}
\par\end{centering}

\protect\protect\protect\caption{An example of $3$ frames of the each movie: (a) the left pendulum,
and (b) the right pendulum.}
\label{fig:Clean_Observations} 
\end{figure}

We generate two movies, each is $5$ seconds long with $N=400$ frames
(namely, sampling interval of $T_{s}=0.0125s$) of each of the pendulums
in the couple pendulum system oscillating in a linear regime as demonstrated
in Figure \ref{fig:Clean_Observations}.

Let $\boldsymbol{m}_{i}^{\left(1\right)}\in\mathbb{R}^{800}$ and
$\boldsymbol{m}_{i}^{\left(2\right)}\in\mathbb{R}^{800}$ be two column
stack vectors consisting of the pixels of the \emph{i}th frame of
the movies of the left and right pendulums, respectively. Let $\mathcal{X}=\left\{ \boldsymbol{x}_{i}\right\} _{i=1}^{N}$
and $\mathcal{Y}=\left\{ \boldsymbol{y}_{i}\right\} _{i=1}^{N}$ be
two sets of observations, which are random projections of the frames
of the movies. In the words, each observation is given by $\boldsymbol{x}_{i}=\boldsymbol{F}\boldsymbol{m}_{i}^{\left(1\right)}$
and $\boldsymbol{y}_{i}=\boldsymbol{G}\boldsymbol{m}_{i}^{\left(2\right)}$,
where $\boldsymbol{F}\in\mathbb{R}^{200\times800}$ and $\boldsymbol{G}\in\mathbb{R}^{200\times800}$
are two fixed matrices with (approximately) orthonormal columns, drawn
independently (once) from a Gaussian distribution. These observations/projections
represent two different modalities in two different spaces. 

We apply Algorithm \prettyref{alg:2} to $\mathcal{X}$ and $\mathcal{Y}$,
where we use $8$ adjacent projected frames (in time), i.e., $\mathcal{X}_{i}=\left\{ \boldsymbol{x}_{j}\right\} _{j=i-3}^{i+4}$
and $\mathcal{Y}_{i}=\left\{ \boldsymbol{y}_{j}\right\} _{j=i-3}^{i+4}$,
as the subset of each observation.

\begin{figure}[t]
\subfloat[]{ \protect\protect\protect\includegraphics[width=1\columnwidth]{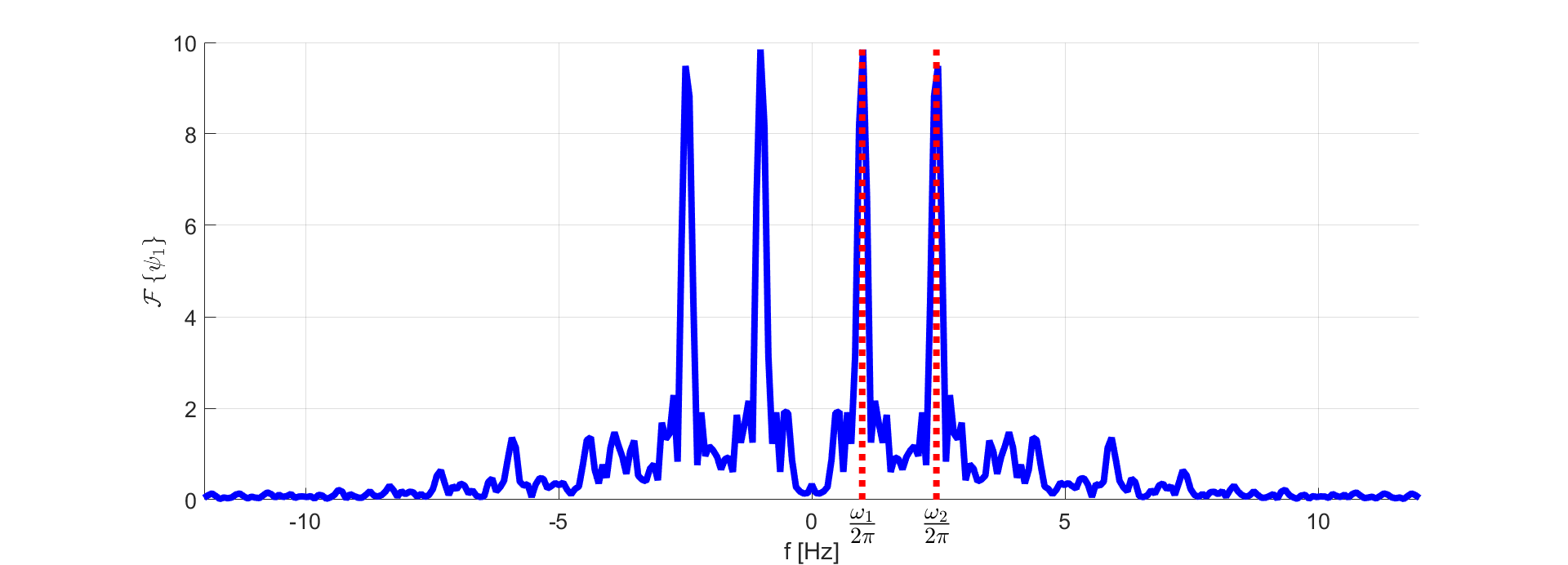}

}

\subfloat[]{\protect\protect\protect\includegraphics[width=1\columnwidth]{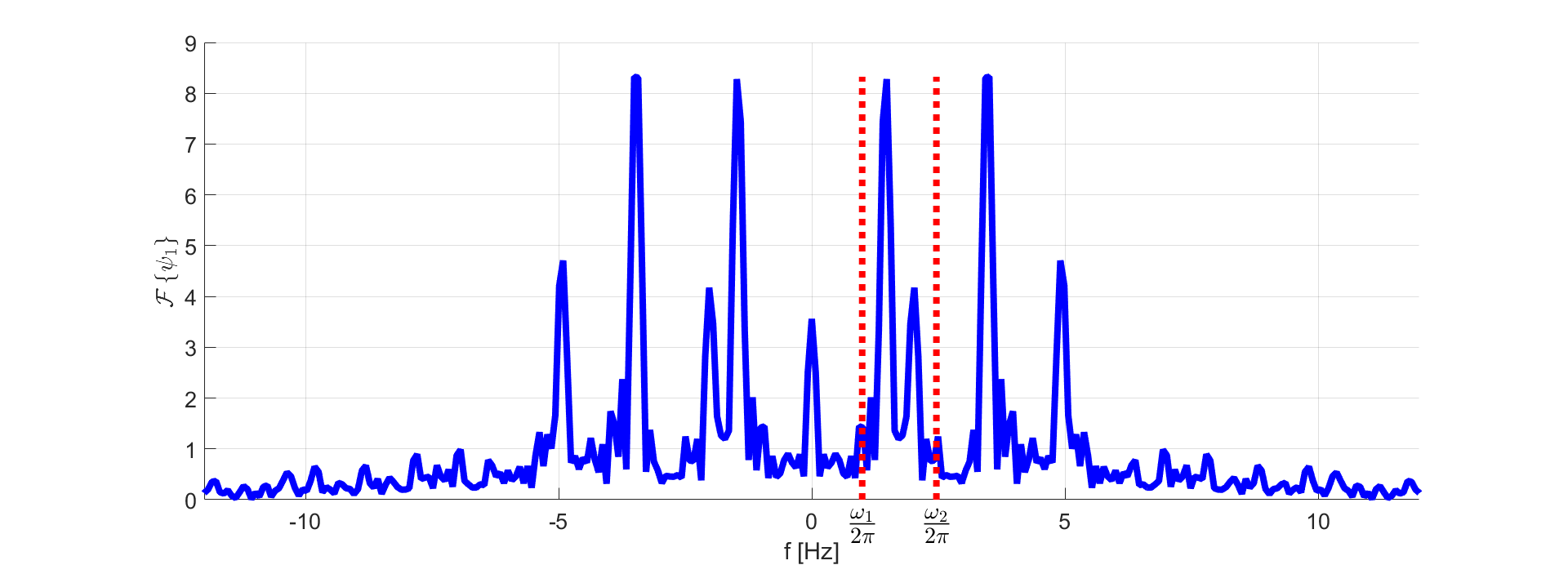}

}

\subfloat[]{\protect\protect\protect\includegraphics[width=1\columnwidth]{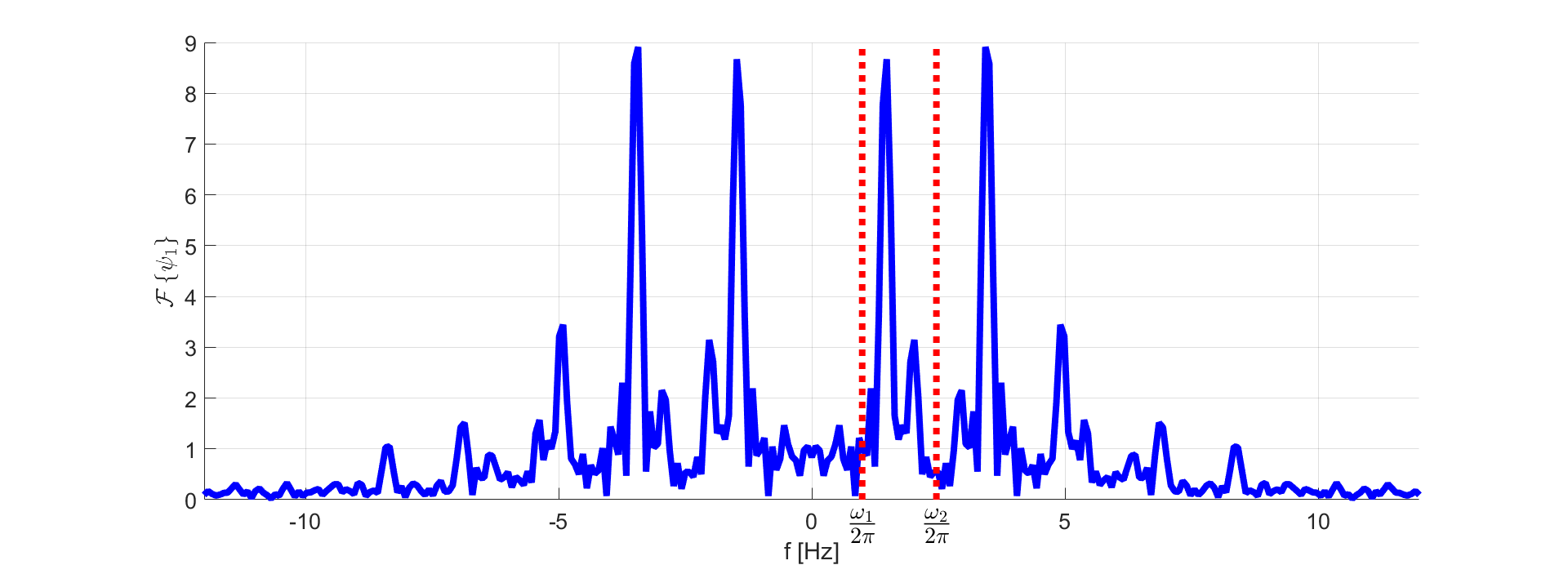}

}

\subfloat[]{\protect\protect\protect\includegraphics[width=1\columnwidth]{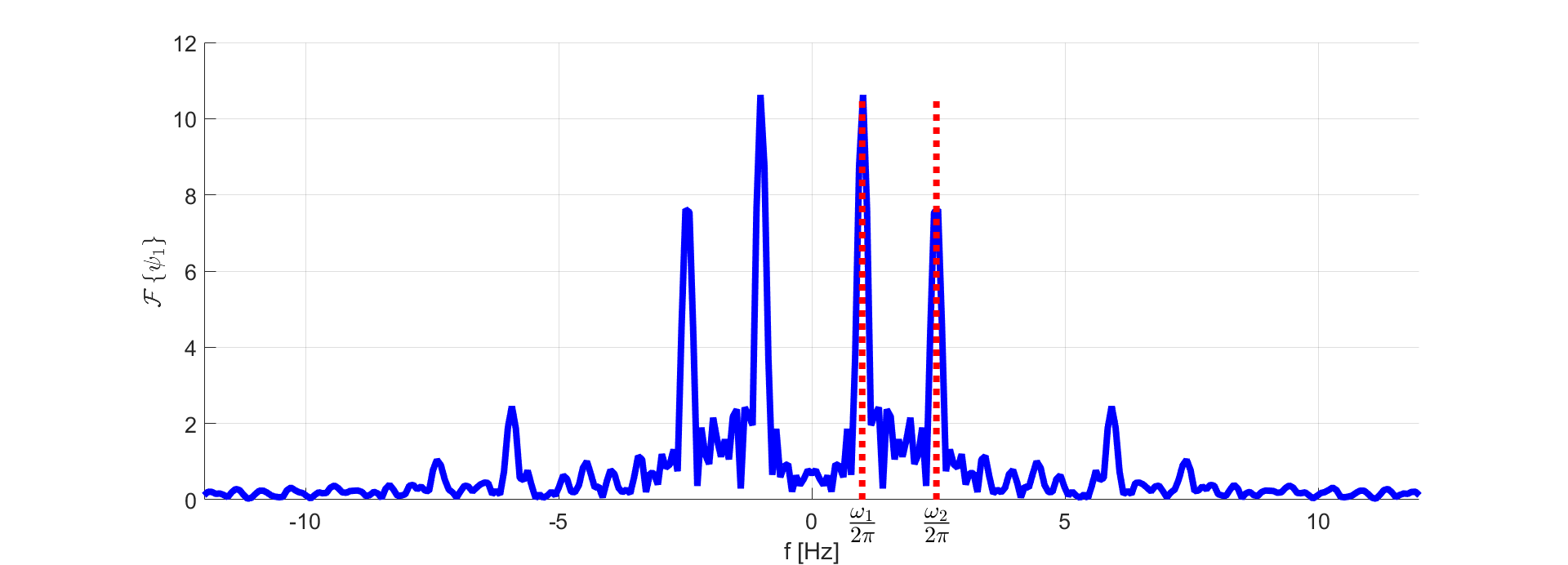}

}

\protect\protect\protect\caption{The Fourier transform of the dominant eigenvectors attained by: (a)
Diffusion Maps based on a single set $\mathcal{X}$, (b) KCCA, (c)
Alternating Diffusion Maps, and (d) Algorithm \prettyref{alg:2}.
The blue curves are the Fourier transforms of the eigenvectors and
the vertical dashed lines represent the two frequencies of the system
$\omega_{1}$ and $\omega_{2}$.}
\label{fig:Clean-movie-Results} 
\end{figure}

We compare our method to $3$ different algorithms: (i) Diffusion
Maps with Euclidean metric (using only $\mathcal{X}$), (ii) KCCA,
and (iii) Alternating Diffusion Maps (with Euclidean metric) \cite{Lederman2015}.
In all four algorithms, we view the nontrivial eigenvector associated
with the largest eigenvalue as the parametrization of the system,
and we display its Fourier transform in Figure~\ref{fig:Clean-movie-Results}.
In each subfigure the blue line is the Fourier transform of the eigenvector
and the two vertical red dashed lines are the two frequencies of the
coupled pendulum system: $\omega_{1}$ and $\omega_{2}$. Figure~\ref{fig:Clean-movie-Results}(a)
displays the output of Diffusion Maps with the Euclidean metric using
the set of observations $\mathcal{X}$ from only one movie. Figure~\ref{fig:Clean-movie-Results}(b)
displays the output of KCCA. Figure~\ref{fig:Clean-movie-Results}(c)
displays the output of Alternating Diffusion Maps. Finally, Figure~\ref{fig:Clean-movie-Results}(d)
displays the output of Algorithm \prettyref{alg:2}, i.e., Diffusion
Maps with $D_{ij}$ \eqref{eq:prop_metric} as its metric. We note
that only one nontrivial eigenvector is presented, since, as shown
in \eqref{eq:ODE_Solution}, the displacement of each pendulum, and
hence, the projected frames of each movie, can be represented by a
single scalar $\theta$, which is the angle of the pendulum with respect
to the equilibrium axis. In other words, the coupled pendulum system
can be described using a low-dimensional representation conveyed by
the dominant nontrivial eigenvector.

As we can see in Figure~\ref{fig:Clean-movie-Results}, both in the
result obtained by Diffusion Maps as well as in the result obtained
by our method, the presented eigenvector contains the frequencies
of the coupled pendulum system: $\omega_{1}$ and $\omega_{2}$. In
contrast, the eigenvector attained by Alternating Diffusion Maps and
the eigenvector attained by KCCA do not contain the true frequencies
of the coupled pendulum system. Specifically, we show in Figure~\ref{fig:Clean-movie-Results}(a)
that the true frequencies can be extracted simply by applying diffusion
maps to one of the observation sets. Consequently, we remark that
this experiment serves only as a reference; it implies that in this
noiseless case each of the sets carries the full information on the
system, and as a result, these frequencies are common to both sets.
In the next section, we introduce noise and show that in the noisy
case our algorithm is essential.

\subsubsection{Case II -- Coupled pendulum with observation-specific noise}

\begin{figure}[t]
\begin{centering}
\subfloat[]{\protect

\protect

\protect\begin{centering}
\protect\protect\protect\includegraphics[width=0.95\columnwidth]{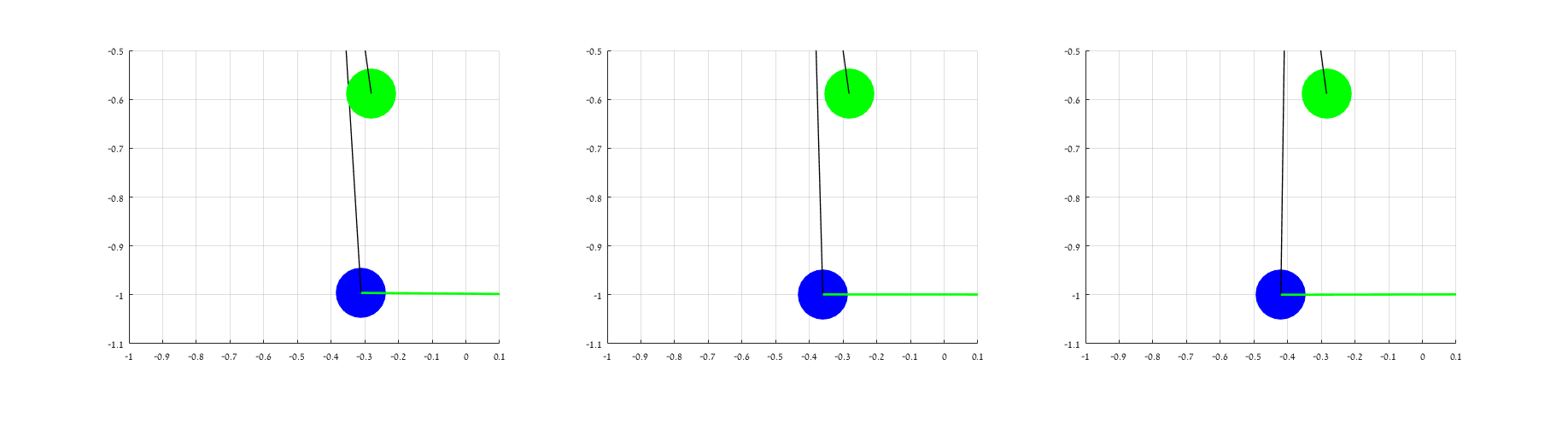}
\protect \protect \protect
\par\end{centering}

}
\par\end{centering}

\begin{centering}
\subfloat[]{\protect

\protect

\protect\begin{centering}
\protect\protect\protect\includegraphics[width=0.95\columnwidth]{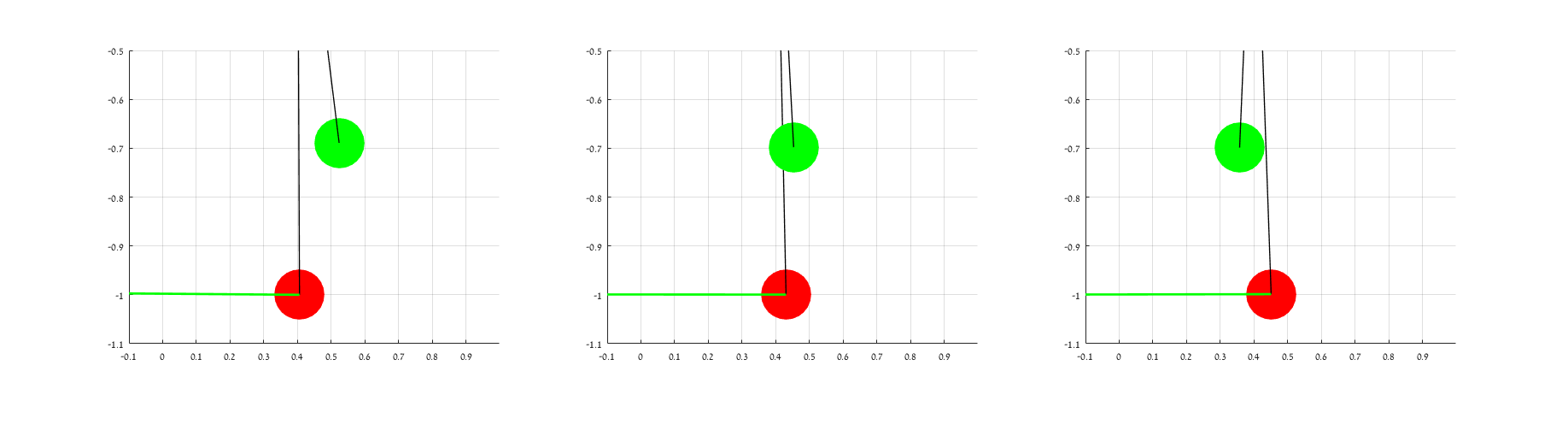}
\protect \protect \protect
\par\end{centering}

}
\par\end{centering}

\protect\protect\protect\caption{An example of $3$ frames of the noisy movie. (a) The left movie which
captures the left coupled pendulum (blue) and an additional simple
pendulum (green). (b) The right movie which captures the right coupled
pendulum (red) and an additional simple pendulum (green).}
\label{fig:Noisy_Movie} 
\end{figure}

We repeat the experiment described in Section \ref{sub:Simulation_Case_I}
with additional observation-specific noise. In this experiment, an
additional simple pendulum is added to each movie as demonstrated
in Figure~\ref{fig:Noisy_Movie}. Note that the two extra pendulums,
one in each movie, oscillate in different frequencies: $\omega_{3}=\frac{1}{5}\omega_{1}$
and $\omega_{4}=4\omega_{1}$. In other words, we now have $4$ different
frequencies in the movies, yet only $2$ of them ($\omega_{1}$ and
$\omega_{2}$) are common to both observations.

We apply Algorithm \ref{alg:2} with the same selection of subsets
$\mathcal{X}_{i}$ and $\mathcal{Y}_{i}$. As in Section \ref{sub:Simulation_Case_I},
we compare our method to the same 3 algorithms.

\begin{figure}[t]
\subfloat[]{ \protect\protect\protect\includegraphics[width=1\columnwidth]{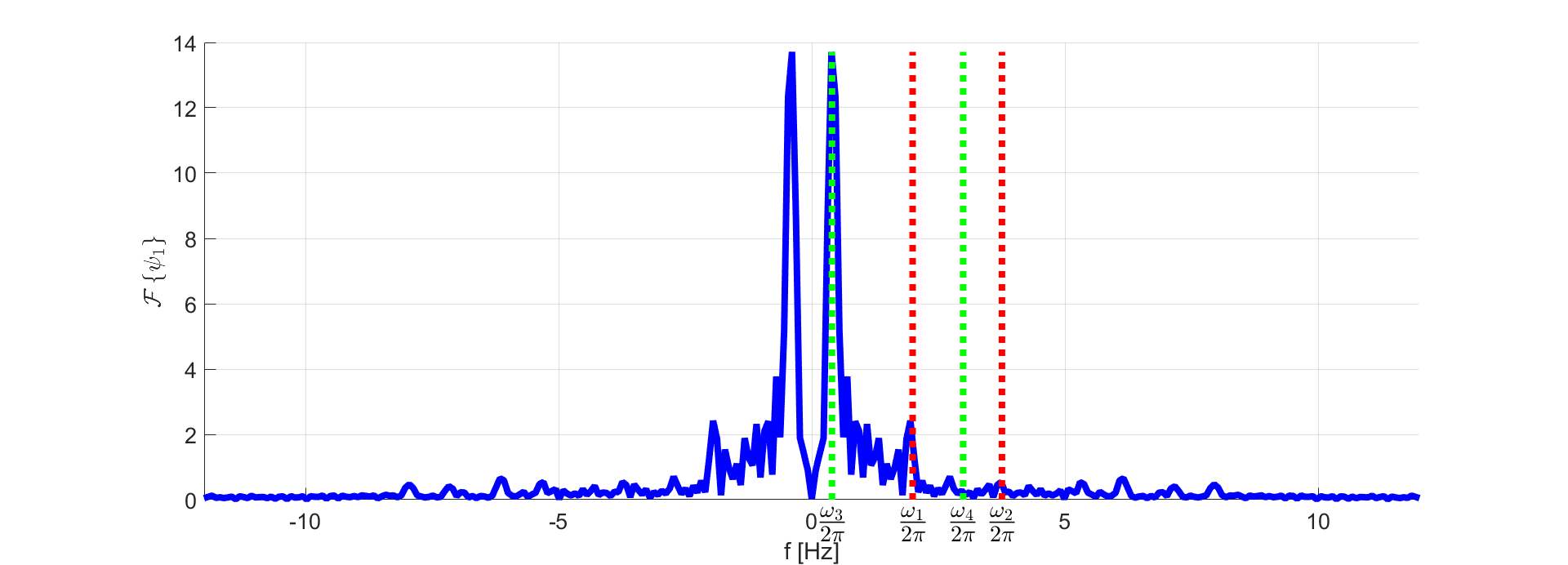}

}

\subfloat[]{ \protect\protect\protect\includegraphics[width=1\columnwidth]{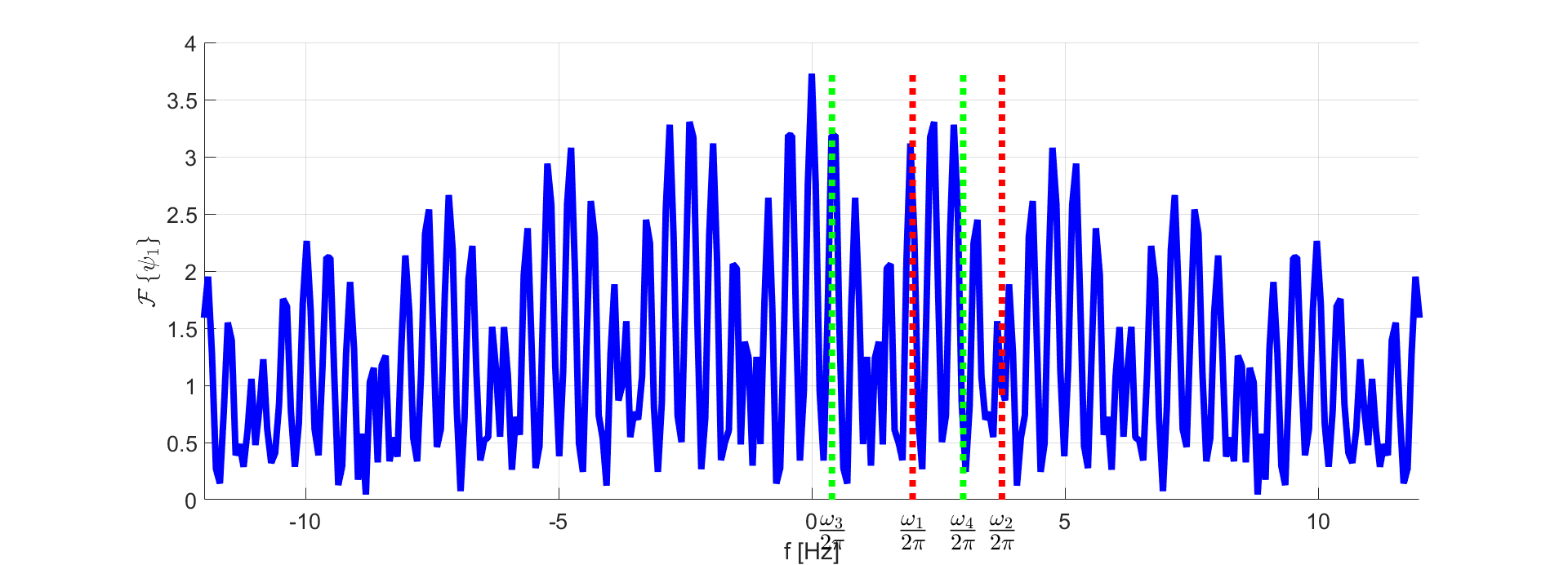}

}

\subfloat[]{ \protect\protect\protect\includegraphics[width=1\columnwidth]{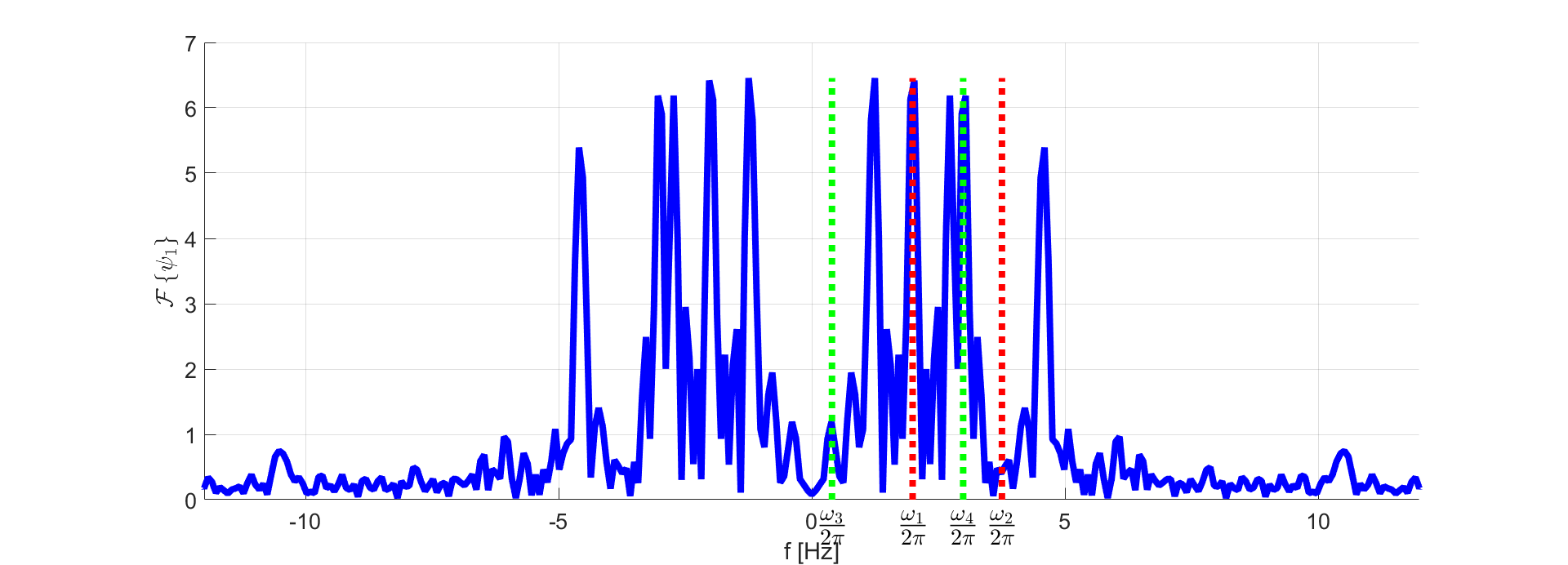}

}

\subfloat[]{ \protect\protect\protect\includegraphics[width=1\columnwidth]{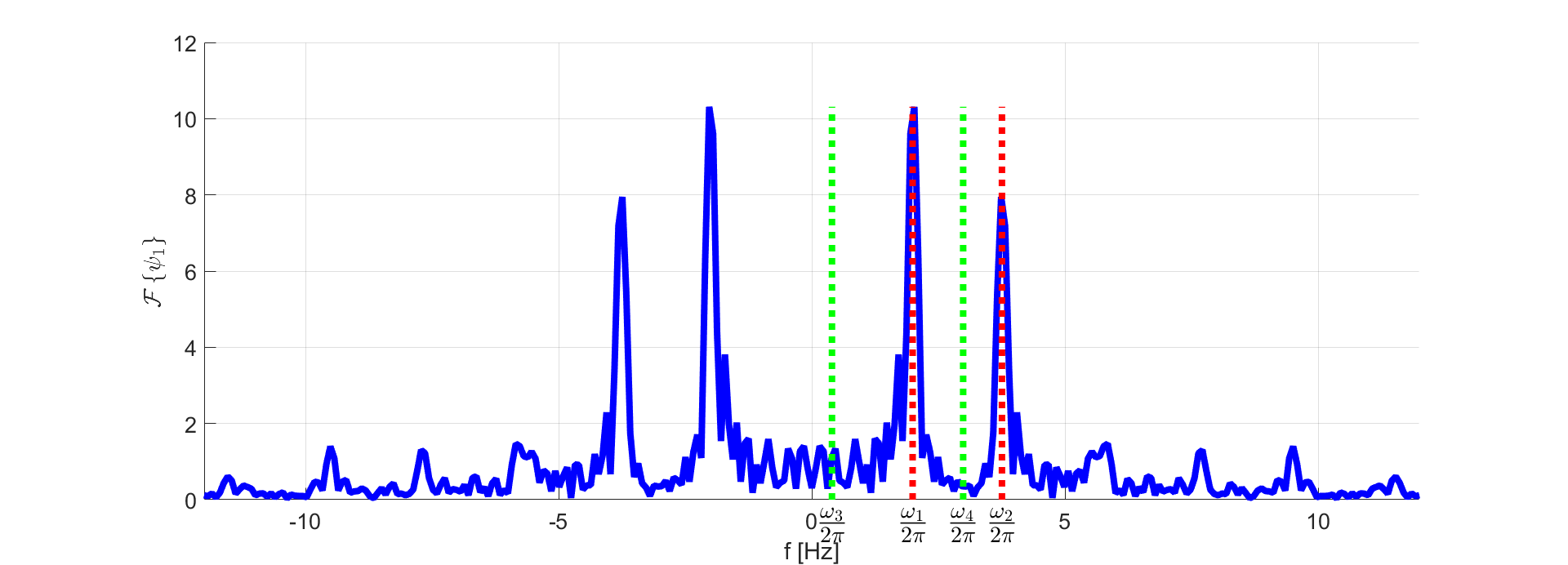}

}

\protect\protect\protect\caption{The output of the $4$ algorithms applied to the coupled pendulum
movies with additional noise: (a) Diffusion Maps, (b) KCCA, (c) Alternating
Diffusion Maps, and (d) Algorithm \ref{alg:2}. The Fourier transforms
of the dominant non-trivial eigenvectors are displayed as blue curves.
The red and green dashed vertical lines represent the frequencies
of the coupled pendulum and the simple uncoupled pendulum, respectively.}
\label{fig:Noisy-movie-Results} 
\end{figure}

Figure~\ref{fig:Noisy-movie-Results} is similar to Figure~\ref{fig:Clean-movie-Results},
where we display the Fourier transforms of the eigenvectors obtained
by the algorithms. In each subfigure the blue curve is the Fourier
transform of the eigenvector, the two red dashed vertical lines are
the two frequencies of the coupled pendulum $\omega_{1}$ and $\omega_{2}$,
and the two green dashed vertical lines are the two frequencies of
the simple pendulums $\omega_{3}$ and $\omega_{4}$. Figure~\ref{fig:Noisy-movie-Results}(a)
displays the output of Diffusion Maps with the Euclidean metric using
frames from only one movie $\mathcal{X}$. Figure~\ref{fig:Noisy-movie-Results}(b)
displays the output of KCCA. Figure~\ref{fig:Noisy-movie-Results}(c)
displays the output of Alternating Diffusion Maps. Finally, Figure~\ref{fig:Noisy-movie-Results}(d)
displays the output of Algorithm \ref{alg:2}, i.e., Diffusion Maps
with the metric $D_{ij}$.

As we can see in Figure~\ref{fig:Noisy-movie-Results}, only in the
result obtained Algorithm \ref{alg:2}, the dominant eigenvector contains
the frequencies of the coupled pendulum system $\omega_{1}$ and $\omega_{2}$
as desired. The eigenvectors attained by Diffusion Maps, by KCAA,
and by Alternating Diffusion Maps do not contain the true frequencies
of the coupled pendulum system.

\subsection{Multiple Observations of Rotating Icons}

\begin{figure}[t]
\begin{centering}
\subfloat[]{\protect

\protect

\protect\begin{centering}
\fbox{%
\begin{minipage}[t]{0.25\columnwidth}%
\protect

\protect

\protect\begin{center}
\protect\protect\protect\includegraphics[width=0.9\columnwidth]{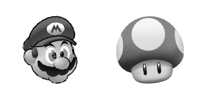}
\protect \protect \protect
\par\end{center}%
\end{minipage}} \protect \protect \protect
\par\end{centering}

}\hfill{}\subfloat[]{\protect

\protect

\protect\begin{centering}
\fbox{%
\begin{minipage}[t]{0.25\columnwidth}%
\protect

\protect

\protect\begin{center}
\protect\protect\protect\includegraphics[width=0.9\columnwidth]{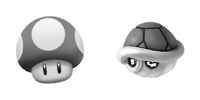}
\protect \protect \protect
\par\end{center}%
\end{minipage}} \protect \protect \protect
\par\end{centering}

}\hfill{}\subfloat[]{\protect

\protect

\protect\begin{centering}
\fbox{%
\begin{minipage}[t]{0.25\columnwidth}%
\protect

\protect

\protect\begin{center}
\protect\protect\protect\includegraphics[width=0.9\columnwidth]{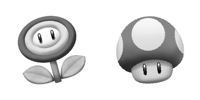}
\protect \protect \protect
\par\end{center}%
\end{minipage}} \protect \protect \protect
\par\end{centering}

}
\par\end{centering}

\begin{centering}
\subfloat[]{\protect

\protect

\protect\begin{centering}
\fbox{%
\begin{minipage}[t]{0.25\columnwidth}%
\protect

\protect

\protect\begin{center}
\protect\protect\protect\includegraphics[width=0.9\columnwidth]{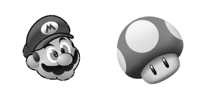}
\protect \protect \protect
\par\end{center}%
\end{minipage}} \protect \protect \protect
\par\end{centering}

}\hfill{}\subfloat[]{\protect

\protect

\protect\begin{centering}
\fbox{%
\begin{minipage}[t]{0.25\columnwidth}%
\protect

\protect

\protect\begin{center}
\protect\protect\protect\includegraphics[width=0.9\columnwidth]{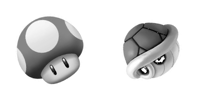}
\protect \protect \protect
\par\end{center}%
\end{minipage}} \protect \protect \protect
\par\end{centering}

}\hfill{}\subfloat[]{\protect

\protect

\protect\begin{centering}
\fbox{%
\begin{minipage}[t]{0.25\columnwidth}%
\protect

\protect

\protect\begin{center}
\protect\protect\protect\includegraphics[width=0.9\columnwidth]{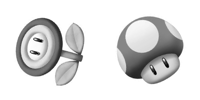}
\protect \protect \protect
\par\end{center}%
\end{minipage}} \protect \protect \protect
\par\end{centering}

}
\par\end{centering}

\protect\protect\protect\caption{Two frames of each of the movies: (a) and (d) are from the first movie,
(b) and (e) from the second movie, and (c) and (f) from the third
movie. Mushroom is common to all the movies while the other icons
are specific for each movie.}
\label{fig:TCCA_Movie} 
\end{figure}

In this simulation we show that Algorithm \ref{alg:full_TCCA} allows
for the accurate parametrization of the common variable underlying
$K=3$ nonlinear high-dimensional observations. We generate three
high-dimensional movies containing four rotating icons: Super Mario,
Mushroom, Turtle and Flower. Each movie captures only two icons. In
the movies, each icon rotates in a constant angular speed: the angular
speeds of Super Mario, Mushroom, Turtle and Flower are $4^{\circ}$,$6^{\circ}$,$10^{\circ}$,
and $15^{\circ}$ per frame, respectively, as demonstrated in Figure~\ref{fig:TCCA_Movie}.
Notice that only Mushroom appears in all the movies, and hence, the
angular speed of Mushroom is the hidden common variable $z$, whereas
the angular speeds of the other icons are the hidden observation-specific
variables $\epsilon^{\left(k\right)}$, for $1\leq k\leq K$. Two
frames of each movie are depicted for illustration in Figure~\ref{fig:TCCA_Movie}.

\begin{figure}[t]
\begin{centering}
\includegraphics[width=0.99\columnwidth]{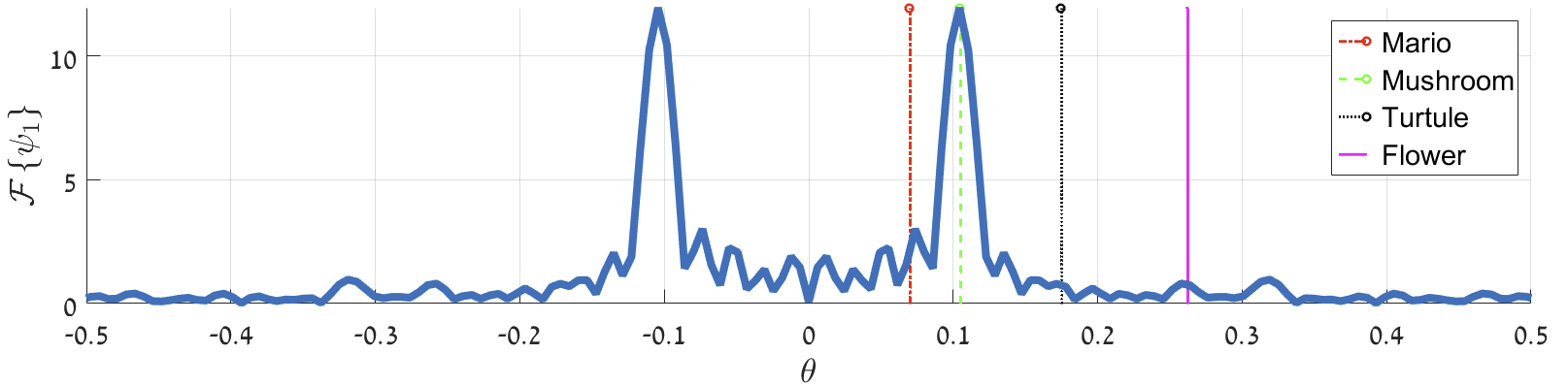} 
\par\end{centering}

\protect\protect\protect\caption{The Fourier transform of the parametrization of the common variable
obtained by Algorithm \ref{alg:full_TCCA}. The vertical dashed lines
represent the true frequencies of the rotating icons.}
\label{fig:TCCA_Result} 
\end{figure}

Each set of nonlinear high-dimensional observations $\mathcal{X}^{\left(k\right)}=\left\{ \boldsymbol{x}_{i}^{\left(k\right)}\right\} _{i=1}^{N}$
consists of $N=300$ frames. We apply Algorithm \ref{alg:full_TCCA},
where in step 1(a), the subsets $\mathcal{X}_{i}^{\left(k\right)}$
consist of the frames in a time window of length $7$ around $\boldsymbol{x}_{i}^{\left(k\right)}$.
Since, the desired parametrization should convey the fact that the
common variable in the sets is the angular speed of Mushroom, and
thus, it should be periodic with the same period, we apply the Fourier
transform to the first column of $\boldsymbol{\Psi}$ and present
it in Figure~\ref{fig:TCCA_Result}. In Figure~\ref{fig:TCCA_Result},
the true frequencies of Super Mario, Mushroom, Turtle and Flower are
marked by vertical red, green, black and pink dashed lines, respectively.

Figure \ref{fig:TCCA_Result} shows that indeed the proposed algorithm
identifies the frequency of Mushroom (the common variable underlying
all observation sets), whereas the frequencies of the observation-specific
Super Mario, Turtle and Flower are completely missing, as expected.

\begin{figure}[t]
\begin{centering}
\subfloat[]{\protect

\protect

\protect\begin{centering}
\fbox{%
\begin{minipage}[t]{0.25\columnwidth}%
\protect

\protect

\protect\begin{center}
\protect\protect\protect\includegraphics[width=0.9\columnwidth]{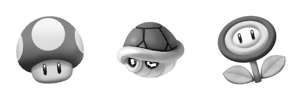}
\protect \protect \protect
\par\end{center}%
\end{minipage}} \protect \protect \protect
\par\end{centering}

}\hfill{}\subfloat[]{\protect

\protect

\protect\begin{centering}
\fbox{%
\begin{minipage}[t]{0.25\columnwidth}%
\protect

\protect

\protect\begin{center}
\protect\protect\protect\includegraphics[width=0.9\columnwidth]{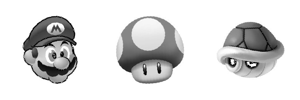}
\protect \protect \protect
\par\end{center}%
\end{minipage}} \protect \protect \protect
\par\end{centering}

}\hfill{}\subfloat[]{\protect

\protect

\protect\begin{centering}
\fbox{%
\begin{minipage}[t]{0.25\columnwidth}%
\protect

\protect

\protect\begin{center}
\protect\protect\protect\includegraphics[width=0.9\columnwidth]{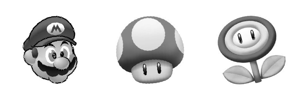}
\protect \protect \protect
\par\end{center}%
\end{minipage}} \protect \protect \protect
\par\end{centering}

}
\par\end{centering}

\begin{centering}
\subfloat[]{\protect

\protect

\protect\begin{centering}
\fbox{%
\begin{minipage}[t]{0.25\columnwidth}%
\protect

\protect

\protect

\protect\begin{center}
\protect\protect\protect\includegraphics[width=0.9\columnwidth]{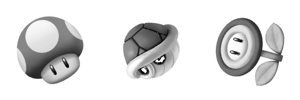}
\protect \protect \protect
\par\end{center}%
\end{minipage}} \protect \protect \protect
\par\end{centering}

}\hfill{}\subfloat[]{\protect

\protect

\protect\begin{centering}
\fbox{%
\begin{minipage}[t]{0.25\columnwidth}%
\protect

\protect

\protect\begin{center}
\protect\protect\protect\includegraphics[width=0.9\columnwidth]{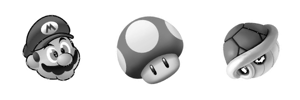}
\protect \protect \protect
\par\end{center}%
\end{minipage}} \protect \protect \protect
\par\end{centering}

}\hfill{}\subfloat[]{\protect

\protect

\protect\begin{centering}
\fbox{%
\begin{minipage}[t]{0.25\columnwidth}%
\protect

\protect

\protect\begin{center}
\protect\protect\protect\includegraphics[width=0.9\columnwidth]{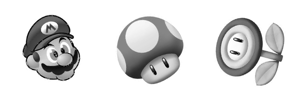}\protect
\protect \protect \protect
\par\end{center}%
\end{minipage}} \protect \protect \protect
\par\end{centering}

}
\par\end{centering}

\protect\protect\protect\caption{Two frames from each movie: (a) and (d) are from the first movie,
(b) and (e) from the second movie, and (c) and (f) from the third
movie. Each two movies contain two common icons. Only Mushroom is
common to all the movies.}
\label{fig:TCCA_Movie_2} 
\end{figure}

To further demonstrate the capabilities of our method, we repeat the
simulation in a more complex setting. Here, each pair of movies contains
two common icons while only Mushroom is maintained as the common variable
of all the movies. Two frames of the new movies are depicted for illustration
in Figure \ref{fig:TCCA_Movie_2}.

\begin{figure}[t]
\begin{centering}
\includegraphics[width=0.99\columnwidth]{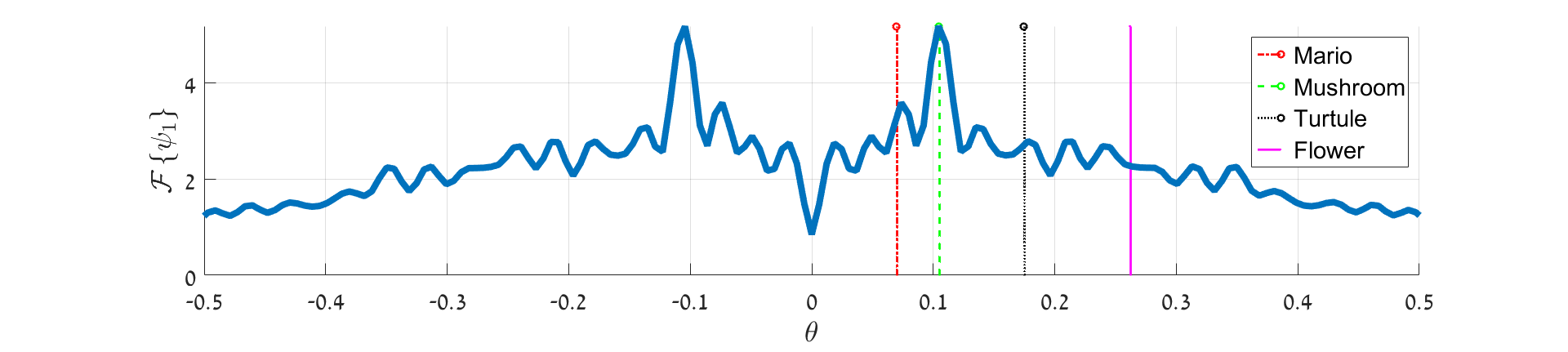} 
\par\end{centering}

\protect\protect\protect\caption{The Fourier transform of the parametrization of the common variable
obtained by Algorithm \ref{alg:full_TCCA}. The vertical dashed lines
represent the true frequencies of the rotating icons.}
\label{fig:TCCA_Result_2} 
\end{figure}

For similar reasons as in the previous experiment, we apply the Fourier
transform to the first column of $\boldsymbol{\Psi}$ and present
it in Figure~\ref{fig:TCCA_Result_2}. In Figure~\ref{fig:TCCA_Result_2},
the true frequencies of Super Mario, Mushroom, Turtle and Flower are
marked by vertical dashed red, green, black and pink lines, respectively.
Despite the more complex setting, in which any two observation sets
contain additional correlated ``noise'', Figure~\ref{fig:TCCA_Result_2}
shows that the proposed algorithm identifies the true frequency of
the common variable (Mushroom).

In summary, in this simulation without assuming any prior knowledge
on the structure and content of the data, our extended method successfully
discovers the common variable hidden in {\em multiple} high-dimensional
and nonlinear observations.

\section{Conclusions \label{sec:Conclusions}}

In this paper, we have presented a new manifold learning method for
extracting the common hidden variables underlying multimodal data
sets. Our method does not assume prior knowledge on the system nor
on the observed data and relies on a local metric, which is learned
from data in an unsupervised manner. Specifically, we proposed a metric
between observations based on local CCA and showed that this metric
approximates the Euclidean distance between the respective hidden
common variables.

The theoretical results were validated in simulations, where we demonstrated
the accurate recovery of the hidden common variables from multiple
complex and high-dimensional data sets. In addition, we showed that
our method can be applied to various different types of observations
and attain the same results without adjusting the algorithm to the
specific observations at hand. For example, the coupled pendulum system
is an example of a dynamical system with a definitive model and a
closed-form solution in the linear case. We have shown that without
any prior model knowledge our method can obtain an accurate description
of the solution solely from high-dimensional nonlinear observations.
Note that our solution was obtained also when the observations contained
``structured noise''.

The capability to obtain the close-form solution solely from observations
enables us to demonstrate the power of our approach by carrying out
empirical modeling of dynamical systems. One can further extend this
to the analysis of the coupled pendulum system in more complex scenarios,
such as, with different initial conditions and in nonlinear regimes.
In such cases, closed-form solutions may no longer be available. Yet,
from a data-driven point of view, our method is expected to attain
an accurate description of the system from its observations. Importantly,
since our method does not require prior rigid model assumptions, it
can be applied to a broad variety of multimodal data sets lacking
definitive models. Therefore, future work will address the extension
of our analysis to various types of dynamical systems and empirical
physics experiments.

 \bibliographystyle{IEEEbib}
\bibliography{refs}

\begin{thebibliography}{10}

\bibitem{lahat2015multimodal}
D.~Lahat, T.~Adali, and C.~Jutten,
\newblock ``Multimodal data fusion: an overview of methods, challenges, and
  prospects,''
\newblock {\em Proc. IEEE}, vol. 103, no. 9, pp. 1449--1477, 2015.

\bibitem{hotelling36cca}
H.~Hotelling,
\newblock ``{Relations Between Two Sets of Variates},''
\newblock {\em Biometrika}, vol. 28, no. 3/4, pp. 321--377, Dec. 1936.

\bibitem{hardoon2004canonical}
D.~R. Hardoon, S.~Szedmak, and J.~Shawe-Taylor,
\newblock ``Canonical correlation analysis: An overview with application to
  learning methods,''
\newblock {\em Neural computation}, vol. 16, no. 12, pp. 2639--2664, 2004.

\bibitem{bach2005probabilistic}
F.~R. Bach and M.~I. Jordan,
\newblock ``A probabilistic interpretation of canonical correlation analysis,''
\newblock 2005.

\bibitem{lai2000kernel}
P.~L. Lai and C.~Fyfe,
\newblock ``Kernel and nonlinear canonical correlation analysis,''
\newblock {\em International Journal of Neural Systems}, vol. 10, no. 5, pp.
  365--377, 2000.

\bibitem{zheng2006facial}
Wenming Zheng, Xiaoyan Zhou, Cairong Zou, and Li~Zhao,
\newblock ``Facial expression recognition using kernel canonical correlation
  analysis (kcca),''
\newblock {\em Neural Networks, IEEE Transactions on}, vol. 17, no. 1, pp.
  233--238, 2006.

\bibitem{melzer2003appearance}
Thomas Melzer, Michael Reiter, and Horst Bischof,
\newblock ``Appearance models based on kernel canonical correlation analysis,''
\newblock {\em Pattern recognition}, vol. 36, no. 9, pp. 1961--1971, 2003.

\bibitem{hardoon2007unsupervised}
David~R Hardoon, Janaina Mourao-Miranda, Michael Brammer, and John
  Shawe-Taylor,
\newblock ``Unsupervised analysis of fmri data using kernel canonical
  correlation,''
\newblock {\em NeuroImage}, vol. 37, no. 4, pp. 1250--1259, 2007.

\bibitem{de2005spectral}
V.~R. de~Sa,
\newblock ``Spectral clustering with two views,''
\newblock in {\em ICML workshop on learning with multiple views}, 2005, pp.
  20--27.

\bibitem{wang2012unsupervised}
Bo~Wang, Jiayan Jiang, Wei Wang, Zhi-Hua Zhou, and Zhuowen Tu,
\newblock ``Unsupervised metric fusion by cross diffusion,''
\newblock in {\em Computer Vision and Pattern Recognition (CVPR), 2012 IEEE
  Conference on}. IEEE, 2012, pp. 2997--3004.

\bibitem{boots2012}
B.~Boots and G.~Gordon,
\newblock ``Two-manifold problems with applications to nonlinear system
  identification,''
\newblock in {\em ICML}, 2012.

\bibitem{Lederman2015}
R.~R. Lederman and R.~Talmon,
\newblock ``Learning the geometry of common latent variables using
  alternating-diffusion,''
\newblock {\em Appl. Comput. Harmon. Anal.}, 2015.

\bibitem{lederman2015alternating}
Roy~R Lederman, Ronen Talmon, Hau-Tieng Wu, Yu-Lun Lo, and Ronald~R Coifman,
\newblock ``Alternating diffusion for common manifold learning with application
  to sleep stage assessment,''
\newblock in {\em Acoustics, Speech and Signal Processing (ICASSP), 2015 IEEE
  International Conference on}. IEEE, 2015, pp. 5758--5762.

\bibitem{roweis2000nonlinear}
S.~T. Roweis and L.~K. Saul,
\newblock ``Nonlinear dimensionality reduction by locally linear embedding,''
\newblock {\em Science}, vol. 290, no. 5500, pp. 2323--2326, 2000.

\bibitem{belkin2003laplacian}
M.~Belkin and P.~Niyogi,
\newblock ``Laplacian eigenmaps for dimensionality reduction and data
  representation,''
\newblock {\em Neural computation}, vol. 15, no. 6, pp. 1373--1396, 2003.

\bibitem{coifman2006diffusion}
R.~R. Coifman and S.~Lafon,
\newblock ``Diffusion maps,''
\newblock {\em Appl. Comput. Harmon. Anal.}, vol. 21, no. 1, pp. 5--30, 2006.

\bibitem{Singer2008226}
A.~Singer and R.~R. Coifman,
\newblock ``Non-linear independent component analysis with diffusion maps,''
\newblock {\em Appl. Comput. Harmon. Anal.}, vol. 25, no. 2, pp. 226 -- 239,
  2008.

\bibitem{talmon2013empirical}
R.~Talmon and R.~R. Coifman,
\newblock ``Empirical intrinsic geometry for nonlinear modeling and time series
  filtering,''
\newblock {\em Proceedings of the National Academy of Sciences}, vol. 110, no.
  31, pp. 12535--12540, 2013.

\bibitem{Talmon2015138}
R.~Talmon and R.~R. Coifman,
\newblock ``Intrinsic modeling of stochastic dynamical systems using empirical
  geometry,''
\newblock {\em Appl. Comput. Harmon. Anal.}, vol. 39, no. 1, pp. 138 -- 160,
  2015.

\bibitem{talmon2015manifold}
Ronen Talmon, St{\'e}phane Mallat, Hitten Zaveri, and Ronald~R Coifman,
\newblock ``Manifold learning for latent variable inference in dynamical
  systems,''
\newblock {\em Signal Processing, IEEE Transactions on}, vol. 63, no. 15, pp.
  3843--3856, 2015.

\bibitem{berry2015local}
T.~Berry and T.~Sauer,
\newblock ``Local kernels and the geometric structure of data,''
\newblock {\em Appl. Comput. Harmon. Anal.}, 2015.

\bibitem{giannakis2015dynamics}
Dimitrios Giannakis,
\newblock ``Dynamics-adapted cone kernels,''
\newblock {\em SIAM Journal on Applied Dynamical Systems}, vol. 14, no. 2, pp.
  556--608, 2015.

\bibitem{de2010multi}
Virginia~R De~Sa, Patrick~W Gallagher, Joshua~M Lewis, and Vicente~L Malave,
\newblock ``Multi-view kernel construction,''
\newblock {\em Machine learning}, vol. 79, no. 1-2, pp. 47--71, 2010.

\bibitem{kumar2011co}
Abhishek Kumar, Piyush Rai, and Hal Daume,
\newblock ``Co-regularized multi-view spectral clustering,''
\newblock in {\em Advances in Neural Information Processing Systems}, 2011, pp.
  1413--1421.

\bibitem{lin2011multiple}
Yen~Yu Lin, Tyng~Luh Liu, and Chiou~Shann Fuh,
\newblock ``Multiple kernel learning for dimensionality reduction,''
\newblock {\em Pattern Analysis and Machine Intelligence, IEEE Transactions
  on}, vol. 33, no. 6, pp. 1147--1160, 2011.

\bibitem{huang2012affinity}
Hsin-Chien Huang, Yung-Yu Chuang, and Chu-Song Chen,
\newblock ``Affinity aggregation for spectral clustering,''
\newblock in {\em Computer Vision and Pattern Recognition (CVPR), 2012 IEEE
  Conference on}. IEEE, 2012, pp. 773--780.

\bibitem{boots2012two}
Byron Boots and Geoff Gordon,
\newblock ``Two-manifold problems with applications to nonlinear system
  identification,''
\newblock {\em arXiv preprint arXiv:1206.4648}, 2012.

\bibitem{lindenbaum2015multiview}
Ofir Lindenbaum, Arie Yeredor, Moshe Salhov, and Amir Averbuch,
\newblock ``Multiview diffusion maps,''
\newblock {\em arXiv preprint arXiv:1508.05550}, 2015.

\bibitem{lindenbaum2015learning}
Ofir Lindenbaum, Arie Yeredor, and Moshe Salhov,
\newblock ``Learning coupled embedding using multiview diffusion maps,''
\newblock in {\em Latent Variable Analysis and Signal Separation}, pp.
  127--134. Springer, 2015.

\bibitem{michaeli2015nonparametric}
Tomer Michaeli, Weiran Wang, and Karen Livescu,
\newblock ``Nonparametric canonical correlation analysis,''
\newblock {\em arXiv preprint arXiv:1511.04839}, 2015.

\bibitem{luo2015tensor}
Yong Luo, Dacheng Tao, Kotagiri Ramamohanarao, Chao Xu, and Yonggang Wen,
\newblock ``Tensor canonical correlation analysis for multi-view dimension
  reduction,''
\newblock {\em Knowledge and Data Engineering, IEEE Transactions on}, vol. 27,
  no. 11, pp. 3111--3124, 2015.

\bibitem{de2000multilinear}
Lieven De~Lathauwer, Bart De~Moor, and Joos Vandewalle,
\newblock ``A multilinear singular value decomposition,''
\newblock {\em SIAM journal on Matrix Analysis and Applications}, vol. 21, no.
  4, pp. 1253--1278, 2000.

\bibitem{yair2016multimodal}
O.~Yair and R.~Talmon,
\newblock ``Multimodal metric learning with local {CCA},''
\newblock {\em submitted}, 2016.

\bibitem{kushnir2012anisotropic}
Dan Kushnir, Ali Haddad, and Ronald~R Coifman,
\newblock ``Anisotropic diffusion on sub-manifolds with application to earth
  structure classification,''
\newblock {\em Applied and Computational Harmonic Analysis}, vol. 32, no. 2,
  pp. 280--294, 2012.

\bibitem{EWERBRING198937}
L.~M. Ewerbring and F.~T. Luk,
\newblock ``Canonical correlations and generalized {SVD}: applications and new
  algorithms,''
\newblock in {\em 32nd Annual Technical Symposium}, 1989, pp. 206--222.

\bibitem{dsilva2015data}
Carmeline~J Dsilva, Ronen Talmon, C~William Gear, Ronald~R Coifman, and
  Ioannis~G Kevrekidis,
\newblock ``Data-driven reduction for multiscale stochastic dynamical
  systems,''
\newblock {\em arXiv preprint arXiv:1501.05195}, 2015.

\bibitem{comon2009tensor}
Pierre Comon, Xavier Luciani, and Andr{\'e}~LF De~Almeida,
\newblock ``Tensor decompositions, alternating least squares and other tales,''
\newblock {\em Journal of Chemometrics}, vol. 23, no. 7-8, pp. 393--405, 2009.

\end{thebibliography}

\end{document}